\documentclass[11pt]{article}
\usepackage[final]{acl}

\usepackage{times}
\usepackage{latexsym}

\usepackage[T1]{fontenc}

\usepackage[utf8]{inputenc}

\usepackage{microtype}

\usepackage{inconsolata}

\usepackage{graphicx}

\usepackage{multirow}
\usepackage{amssymb}
\usepackage{subcaption}
\usepackage{booktabs}
\usepackage{amsmath}
\usepackage{cleveref}

\usepackage{makecell}
\usepackage{mathtools}
\usepackage{cuted}
\usepackage{caption}  

\usepackage{booktabs}
\usepackage{arydshln}  
\usepackage{multicol}
\usepackage{array}

\title{Language Proficiency Assessment from Eye Movements \\ in Naturalistic Passage Reading}

\author{Shachar Frenkel$^{*}$, Ido Falah\thanks{Equal contribution.}, Omer Shubi, Yevgeni Berzak \\
 Faculty of Data and Decision Sciences,\\
 Technion - Israel Institute of Technology\\
 \texttt{\{fshachar,ido.falah,shubi\}@campus.technion.ac.il},
\texttt{berzak@technion.ac.il} \\
}

\begin{document}
\maketitle
\begin{abstract}
Standard language proficiency tests rely on linguistic tasks such as vocabulary, grammar and reading comprehension quizzes. An alternative, cognitively motivated approach, introduced in \citet{berzak-etal-2018-assessing}, proposed instead to predict language proficiency from behavioral traces of eye movements in reading. In this work, we validate and extend this approach from single sentences to more naturalistic reading of contextualized passages in English as a second language, new proficiency measures, prediction models, and reading in an information seeking regime.  We find that the approach is effective in all these evaluations. We further address two key open questions on eye movement based proficiency testing: (1) potential scoring biases that reflect the proximity of the reader's native language to English, which may undermine validity, and (2) its reliability. We find that eye movement based proficiency scores are indeed biased towards L1s that are linguistically closer to English. We propose a score debiasing method which effectively remedies this issue. The reliability analyses suggest that eye movement proficiency scores are more reliable than standard language proficiency scores. Overall, our results strengthen and broaden the empirical foundations for future eye movement based language assessment technologies.\footnote{Code is available at \href{https://github.com/lacclab/Proficiency-Assessment-from-Eye-Movements}{github.com/lacclab/Proficiency-Assessment-from-Eye-Movements}.}

\end{abstract}

\section{Introduction}

English has been the de-facto global lingua franca for many decades, and the majority of English speakers worldwide learn, speak and read in English as a foreign language. As a result, educational institutions, English teachers and English learners are in an ever-growing need for effective and accurate tools for assessing English proficiency. Thus far, the highest quality language assessments for English L2 speakers have been developed by public organizations and commercial companies such as the Educational Testing Service (ETS) and Cambridge Assessment English. 

Such assessments are extremely labor-intensive, costly and time-consuming to create. Consequently, they are not deployed frequently, and their commercial variants involve substantial fees for test takers. Furthermore, materials from standardized tests are typically not made publicly available. The scarcity of high-quality materials for language assessments in the public domain leads many institutions and educators to create their own in-house assessments. Those assessments are not only difficult to develop, but are also often of lower quality, as they rarely receive the same attention to item construction, scaling and validation as in standardized tests. While in recent years new possibilities have emerged for faster development of test materials using generative AI, such tests are similarly subject to various quality assurance issues \citep{chuang2025}. 

Importantly, in addition to practical test construction and validation bottlenecks, the testing methodology of standard language proficiency tests itself is fundamentally limited. First, it relies on small samples of ad-hoc test items. Even more importantly, it is based on an \emph{offline signal}: the end responses to these items. It has no ability to trace \emph{online} cognitive processes during language processing as they unfold in real time. 
This limits the informativeness of the test and the quality of the feedback that can be provided to test takers. 

An alternative approach, introduced in \citet{berzak-etal-2018-assessing}, proposed assessing L2 proficiency as a byproduct of ordinary reading, using behavioral traces of eye movements. This approach measures proficiency via \emph{online} cognitive signals, and has the potential of obviating many of the practical drawbacks of traditional proficiency tests. \citet{berzak-etal-2018-assessing} introduced two variants of this framework. The first is EyeScore, a standalone measure which estimates L2 proficiency using the similarity of the reader's eye movements to L1 readers. The second uses eye movement features to predict performance on external standardized proficiency tests. 

While \citet{berzak-etal-2018-assessing} provided  conceptual and methodological frameworks for eye movement based language proficiency testing, the generality and robustness of their approach has not been studied beyond their data, in which participants read single sentences. Furthermore, two key open questions that are pertinent to the real-world adoption potential of eye movement based proficiency tests have not been examined. The first is whether the validity of eye movement based scores may be undermined by biases stemming from the linguistic proximity of the reader's L1 to English. The second question concerns the reliability of such scores, which is yet to be characterized. 

Our primary contributions with respect to these questions are the following.
\begin{itemize}
    \item \textbf{Replication} we validate eye movement based English proficiency scoring on 1,265 L2 participants from two passage reading corpora, MECO L2 \citep{mecoL2-2023,mecoL2-2025} and OneStopL2, using different language proficiency measures, several prediction models, and in two reading regimes: reading for comprehension and information seeking.
    \item \textbf{L1 Bias} we find that EyeScore favors L2 speakers whose native languages are more similar to English. We propose a debiasing method which alleviates this issue.
    \item \textbf{Reliability} we analyze the internal consistency and split-half reliability of EyeScore. We find that EyeScore is more reliable than scores of standard language proficiency tests.
\end{itemize}

\section{Background and Related Work}

The prediction of language proficiency from eye movements is rooted in extensive literature in psycholinguistics which suggests that eye movements are influenced by linguistic properties of the text, and reflect real time language comprehension processes of the reader \citep[e.g.,][]{just1980theory,reichle1998toward,engbert2005swift,demberg2008data,rayner2012psychology}. Additionally, ample evidence indicates that different levels of language proficiency yield differences in real time processing of the linguistic input \citep[e.g.][]{CLAHSEN_FELSER_2006,whitford2012second,cop2015eye,berzak-levy2023}. Combining these two factors, one can posit a causal chain where differences in language proficiency lead to differences in online comprehension processes, which in turn have behavioral traces in eye movements in reading. Decoding linguistic proficiency from eye movements can thus be viewed as an attempt to reverse-engineer this process, starting from the eye movement signal, and inferring back the linguistic proficiency of the reader.

Our work primarily builds on \citet{berzak-etal-2018-assessing} who demonstrated the feasibility of predicting English proficiency from eye movements in reading using single sentence data from CELER \cite{berzak2022celer}, with the Michigan Placement Test and TOEFL as external references. 
Our investigation of validity with respect to the reader's L1 is motivated by work on systematic influences of the L1 on L2 learning and processing \citep{odlin1989language,jarvis2008crosslinguistic,jeon2014l2}. Such influences may impact various aspects of L2 production and comprehension, including reading, %
where prior work has demonstrated that the speaker's L1 can be decoded from eye movement patterns in reading English L2 \cite{berzak2017predicting,reich2022inferring,skerath2023native}.

The current work is further related to studies that used eye movements for the prediction of reading comprehension \citep[e.g.,][]{copeland2014, Ahn2020TowardsBehavior,reich2022inferring,meziere2023using,shubi2024fine}. In these studies, reading comprehension performance is evaluated on the same textual materials for which the eye tracking data is collected. While the two tasks are related, our focus here is on \emph{external} proficiency tests as measures of a more general linguistic ability. 

More broadly, our study is situated within a nascent line of research on using eye movements for prediction of reader characteristics, such as dyslexia \cite[e.g.,][]{rello2015detecting,haller2022dyslexia,laurinavichyute2025automatic}, and reader-text interactions such as different reading regimes \cite[e.g.,][]{shubi2025decoding,meiri2025deja}. EyeBench \cite{shubi2025eyebench} is a recent benchmark which assembles several such tasks. 

\section{Eye Movement Based Estimation of Language Proficiency}

We study two approaches to L2 proficiency testing using eye movements in reading, EyeScore and prediction of external proficiency scores. 

\subsection{EyeScore} 
\label{sec:eyescore-def}

The first type of tests is derived entirely using eye movements in reading, without external references. Specifically, we evaluate EyeScore, proposed by \citet{berzak-etal-2018-assessing}, which determines L2 proficiency via the similarity of eye movement patterns to L1 speakers. This approach is supported by empirical evidence on differences between L1 and L2 readers with respect to standard eye movement measures  \citep{cop2015eye,berzak2022celer,mecoL2-2023} and the extent to which they depend on the text \citep{cop2015frequency,whitford2012second,whitford2017effects,berzak-levy2023}.

The procedure for calculating EyeScore is as follows. Given a set of participants $D$ consisting of L1 readers $D_{L1}$ and L2 readers $D_{L2}$, each participant $p \in D$ is represented by an eye movement feature vector $v_p \in \mathbb{R}^d$, where the dimensionality $d$ depends on the feature set used (see \Cref{sec:representations}). The vectors are then z-scored feature-wise using a z-scaler fitted on  $D_{L2}$, yielding a normalized vector $\tilde{v}_p$ for each participant. The L1 prototype vector $\bar{v}_{L1}$ is the average of the normalized L1 vectors:
$\bar{v}_{L1} = \frac{1}{|D_{L1}|} \sum_{p \in D_{L1}} \tilde{v}_p$.
The EyeScore of an L2 participant is the cosine similarity between their normalized feature vector and the L1 prototype:
\begin{equation} 
\mathrm{EyeScore}_p = \frac{\tilde{v}_p \cdot \bar{v}_{L1}}{\|\tilde{v}_p\| \, \|\bar{v}_{L1}\|}
\end{equation}

\subsection{Predicting Scores on External Tests} The second approach to eye movement based proficiency testing uses eye movements to predict outcomes on standard language proficiency tests. As such, it enables obtaining a performance estimate on an external test of interest, without the need for administering it. Here, we examine using eye movement feature vectors with \textbf{Ridge regression}, as in \citet{berzak-etal-2018-assessing}. We additionally evaluate the decision tree-based model LightGBM~\cite{ke2017lightgbm}, and a recent neural  model for tabular data, TabPFN-3~\cite{grinsztajn2026tabpfn3technicalreport}. While more complex neural models for eye movements and text were introduced in recent years, we focus on the above models, and specifically on Ridge regression, for comparability with \citet{berzak-etal-2018-assessing}, and as standard machine learning models were found to be better suited for eye movement settings with a limited number of samples \cite{shubi2025eyebench}. 

\subsection{Baselines}
\paragraph{Reading Speed (WPM)} Our primary baseline is reading speed, the number of words read per minute (WPM). It captures the tendency of more proficient individuals to read faster. Importantly, it can be obtained without eye tracking equipment, and thus allows assessing the added value of eye movements for proficiency evaluation. In comparisons to EyeScore, we z-normalize WPM as EyeScore features. In comparisons to prediction of external proficiency tests, we use WPM as input to the prediction model.

\paragraph{Average Train Score (Avg. Train)}
For the task of predicting scores on external tests, we further provide a baseline where we assign the score of a test participant to be the mean score among the training set participants.

\section{Eye Movement Representations}
\label{sec:representations}

The eye movement trajectory in reading is divided into fixations, where the gaze position is stable, and saccades, which are rapid transitions between fixations \cite{schotter2025beginner}. 
Our eye movement representations are derived from this trajectory, and largely follow \citet{berzak-etal-2018-assessing}. We describe the feature sets in brief below, and provide additional details in \Cref{sec:app-Eye-Movement-Representations}. 

\paragraph{Average Fixation Metrics (Avg. Fix.)} Fixation and saccade measures: First Fixation Duration (FF), Gaze Duration (GD), Total Fixation Duration (TF) and Go-Past Duration (GP), averaged across all the words, as well as first pass Skip Rate (SR) and first pass Regression Probability (RP). This feature set was not used in \citet{berzak-etal-2018-assessing}.

\paragraph{Syntactic Clusters (S-Clusters)} The above fixation measures averaged per part-of-speech category, using Universal part-of-speech tags \citep{petrov2012universal} and Penn Treebank (PTB) part-of-speech tags \citep{santorini1990part}.

\paragraph{Word Property Coefficients (WP-Coefs)} The linguistic word properties length, frequency and surprisal are known to affect reading times \citep[e.g.,][]{kliegl2004length, rayner2004effects}. We encode the strength of this influence using the coefficients of linear models fitted to predict the fixation measures of each reader from length, frequency and surprisal. Surprisals are obtained from GPT-2 \citep{radford_language_2019}, and frequencies from Wordfreq \citep{robyn_speer_2018}. 

\paragraph{Transitions} The number of saccades in each direction between each pair of words in the text. 

\paragraph{Word Fixation Metrics (Word Fix.)} GD and TF for each individual word token in the text. 
    
We note that the last two feature sets require the same texts for all participants (Fixed Text regime in \Cref{sec:fixed-any}), while the other three are also applicable when different participants read different texts (Any Text regime in \Cref{sec:fixed-any}).

\section{Experimental Setup}

\subsection{Eye Movement Data}

Our study is enabled by three recent large scale eye tracking datasets with passage reading in English (see \Cref{sec:app-data} for additional details). 

\paragraph{MECO L2} \cite{mecoL2-2023,mecoL2-2025} was collected across multiple sites using  EyeLink 1000 (1000Hz) and similar eye trackers. It includes 1,106 L2 participants from 18 L1 backgrounds, and 95 L1 participants. All the participants read the same 12 informational texts (one per screen), and answered two comprehension questions after each text. For any given participant, eye movement data is available for 5--12 of the texts. 

\paragraph{OneStopL2} was collected with an EyeLink 1000 Plus (1000Hz), and has 278 L2 participants from 10 L1 backgrounds. The texts are 30 newswire articles. Each participant reads one of three 10-article batches with 54 passages. Each passage is presented either in its original or simplified form, and is followed by a comprehension question. 154 participants read in an ordinary reading for comprehension regime, and 124 in an information seeking regime, where they receive the question prior to reading the passage. Our main analysis focuses on the reading for comprehension part of the data. 

\paragraph{OneStop} \cite{onestop2025} was collected with an EyeLink 1000 Plus, and includes 360 English L1 participants (180 reading for comprehension, 180 information seeking). It has the same experimental design and the same textual materials as OneStopL2. We use this dataset as an L1 reference for analyses with OneStopL2.

\subsection{Standard English Proficiency Tests}
\label{sec:proficiency-tests}

In addition to eye movements data, the eye tracking datasets above include results for the following proficiency tests taken by the participants in-lab. 

\paragraph{LexTALE} \citep{lemhofer2012introducing} An English vocabulary test which is often used as a proxy for English proficiency. The score range for LexTALE is 0 to 100. Test results are available for all the OneStopL2 participants, 100 OneStop participants, 1,007 L2 participants of MECO L2 and 93 L1 participants of MECO L2. L1 participants with missing LexTALE scores are assigned the average L1 score in each dataset.

\paragraph{Michigan Placement Test} (Form B) A standardized English proficiency test with four sections: listening comprehension, grammar, vocabulary and reading comprehension. It comprises 100 questions, with a score range of 0 to 100. Test results are available for all the OneStopL2 participants. \citet{berzak-etal-2018-assessing} used the listening and grammar sections of this test. We follow them in assigning L1 participants with the maximal test score.

\paragraph{Composite Proficiency} While MECO L2 does not include a standardized English proficiency test, it has the following four tests which are related to language proficiency: spelling recognition  \citep{andrews2010lexical}, vocabulary \citep[][]{beglar2010rasch} (groups 2--5), and two variants (word and pseudoword) of the Test of Word Reading Efficiency (TOWRE) \citep{torgesen1999towre}. We use them to derive a Composite Proficiency score by applying PCA on the participant $\times$ test score matrix, and extracting the first principal component.

\subsection{Prior Eye Movement Data Availability}
\label{sec:fixed-any}

When designing an eye movement based proficiency test, one needs to consider the evaluation setting with respect to which eye movement data is available, from which participants, and on which textual materials. Here, we assume that \emph{no prior eye movement data is available for the test participant}. For the textual materials of the test, we follow \citet{berzak-etal-2018-assessing} in distinguishing between the Fixed Text and Any Text regimes.

\paragraph{Fixed Text} Eye movement data for the textual materials of the test is available from other participants. We evaluate a stringent variant of this setting where no eye movement data is available for texts other than those presented to the test participant. This setting is akin to traditional proficiency tests that rely on specific test forms.

\paragraph{Any Text} No eye movement data is available for the test materials. In other words, all existing eye movement data is on different texts than those presented to the test participant. This is a much more general setting that supports eye tracking based testing using arbitrary texts. 

\subsection{Evaluation Measures}

\textbf{EyeScore} 
The evaluation of standalone eye tracking based proficiency scores is a challenging methodological problem. Following \citet{berzak-etal-2018-assessing}, we use scores from external proficiency tests (LexTALE, Michigan and Composite scores described in \Cref{sec:proficiency-tests}) as heuristic measures for evaluating the \emph{validity} of EyeScore. In \Cref{sec:eyescore-eval,sec:debiasing-validity} we report the Pearson $r$ coefficient between EyeScore and these tests. In \Cref{sec:bias} we examine another aspect of EyeScore's validity by developing a measure of L1  bias relative to standard tests. Cronbach's $\alpha$ and split-half  Pearson $r$ are used for estimating the \emph{reliability} of EyeScore in \Cref{sec:reliability}. 

\paragraph{Direct Prediction of External Scores} 
In \Cref{sec:prediction}, we evaluate the regression models trained to predict LexTALE, Michigan and Composite scores using Pearson $r$ and Mean Absolute Error (MAE) between predicted and true scores. 

\subsection{Data Splits}
\label{sec:eval-setting}

Both EyeScore and the prediction of external test scores are evaluated using \emph{cross validation}. As we focus on proficiency assessment for L2, only L2 participants appear in the test set. For predicting external test scores, we follow \citet{berzak-etal-2018-assessing} in including L1 participants in the training set.

\textbf{MECO L2} We divide the 12-passage dataset into two parts with 6 passages each and use leave-one-participant-out cross validation. We perform the evaluation for each participant twice: once for each half of the data, and average the results. In the Fixed Text regime, the training passages are from the same half of the data read by the test participant, and in the Any Text regime they are from the other half. As not all participants have eye movement data for all 12 passages, in the Fixed Text regime, we impute the missing passage features for the Transitions and the Word Fixation Metrics feature sets with the mean values in the training set. On average, there are 1,148 participants (1,053 L2, 95 L1) in the training set in both Text regimes.

\textbf{OneStopL2} Each participant reads one of three 10-article batches. As there are two variants of each batch with respect to the difficulty version of each passage (original or simplified), there are 6 subgroups of participants reading exactly the same texts. In the Fixed Text regime, we perform leave-one-participant-out cross validation for each group, with an average of 24 L2 and 24 L1 training participants. In the Any Text regime we perform 3-fold cross validation where test participants read one batch, and training participants are from the other two batches, with an average of 100 L2 and 99 L1 training participants. Additional details on the data splits are described in \Cref{sec:app-eval-settings}.

\subsection{Hyperparameter Tuning}
For predicting external proficiency scores with Ridge regression, we tune the Ridge regularization parameter on the training set with cross validation, over 10 log-spaced values between $10^{-3}$ and $10^{4}$. \Cref{sec:app-tuning} provides additional details on this procedure, and on the hyperparameter tuning for LightGBM and TabPFN-3.

\section{Replication Results}
 
\subsection{EyeScore}
\label{sec:eyescore-eval}

\Cref{tab:eyescore_seen_unseen_bootstrap} presents the Pearson $r$ of EyeScore with external English proficiency tests for L2 participants. The results are consistent across the two datasets and three external reference tests, and the general trends are in line with \citet{berzak-etal-2018-assessing}, with mostly similar or higher correlations. Among the three feature sets that apply to both the Fixed and Any Text regimes, WP-Coefs tends to perform best. Importantly, all three feature sets yield comparable results in the Fixed and Any Text regimes, suggesting that they are robust to the specifics of the textual input. The Word Fixation Metrics feature set is the strongest in the Fixed Text regime. 

\begin{table}[ht!]
\centering
\resizebox{\columnwidth}{!}{
\begin{tabular}{lcc|cc|cc|cc}
\toprule
\multicolumn{1}{c}{} & \multicolumn{4}{c|}{\textbf{MECO L2}} & \multicolumn{4}{c}{\textbf{OneStopL2}} \\
\multicolumn{1}{c}{} & \multicolumn{2}{c}{LexTALE} & \multicolumn{2}{c|}{Composite} & \multicolumn{2}{c}{LexTALE} & \multicolumn{2}{c}{Michigan} \\ \cmidrule{2-9}
\textbf{Features} & Fixed & Any & Fixed & Any & Fixed & Any & Fixed & Any \\
\midrule
WPM & $0.48\phantom{^{***}}$ & $0.48\phantom{^{***}}$ & $0.45\phantom{^{***}}$ & $0.45\phantom{^{***}}$ & $0.47\phantom{^{**}}$ & $0.48$ & $0.47\phantom{^{**}}$ & $0.47$ \\
\midrule
Avg. Fix. & $0.49\phantom{^{***}}$ & $0.50\phantom{^{***}}$ & $0.45\phantom{^{***}}$ & $0.46\phantom{^{***}}$ & $0.48\phantom{^{**}}$ & $0.49$ & $0.50\phantom{^{**}}$ & $0.49$ \\
S-Clusters & $0.50\phantom{^{***}}$ & $0.50\phantom{^{***}}$ & $0.47\phantom{^{***}}$ & $0.46\phantom{^{***}}$ & $0.50\phantom{^{**}}$ & $\mathbf{0.51}$ & $0.50\phantom{^{**}}$ & $0.51$ \\
WP-Coefs & $\mathbf{0.54}^{***}$ & $\mathbf{0.54}^{***}$ & $\mathbf{0.51}^{***}$ & $\mathbf{0.51}^{***}$ & $0.49\phantom{^{**}}$ & $\mathbf{0.51}$ & $0.51\phantom{^{**}}$ & $\mathbf{0.52}$ \\
\midrule
Transitions & $0.48\phantom{^{***}}$ & - & $0.40^{*}\phantom{^{**}}$ & - & $0.51\phantom{^{**}}$ & - & $0.49\phantom{^{**}}$ & - \\
Word Fix. & $\mathbf{0.54}^{***}$ & - & $\mathbf{0.51}^{***}$ & - & $\mathbf{0.54}^{**}$ & - & $\mathbf{0.55}^{**}$ & - \\
\bottomrule
\end{tabular}
}
\caption{\textbf{EyeScore}. Pearson $r$ correlations between EyeScore and standard English proficiency tests. Presented are mean $r$ correlations using bootstrap. Statistically significant differences from the WPM baseline using a bootstrap test are marked with `$^{*}$' $p<0.05$, `$^{**}$' $p<0.01$, `$^{***}$' $p<0.001$. ``Fixed'' is the Fixed Text regime in which all the eye movement data is for the same texts presented to the test participant. ``Any'' is the Any Text regime in which no eye movement data is available for the texts of the test participant. The best result for each eye tracking dataset, benchmark proficiency test, and Text regime is marked in bold.}
\label{tab:eyescore_seen_unseen_bootstrap}
\end{table}

\begin{table*}[ht!]
\makeatletter
\@ifundefined{hdashline}
  {\providecommand{\meanbaselinerule}{\addlinespace[2pt]\midrule\addlinespace[2pt]}}
  {\@ifundefined{dashlinedash}{}{\setlength\dashlinedash{2pt}\setlength\dashlinegap{2pt}}%
   \providecommand{\meanbaselinerule}{\addlinespace[2pt]\hdashline\addlinespace[2pt]}}
\makeatother
\centering
\resizebox{\textwidth}{!}{
\begin{tabular}{lc@{\hspace{4pt}}c|c@{\hspace{4pt}}c|c@{\hspace{4pt}}c|c@{\hspace{4pt}}c|c@{\hspace{4pt}}c|c@{\hspace{4pt}}c|c@{\hspace{4pt}}c|c@{\hspace{4pt}}c}
\toprule
\multicolumn{1}{l}{} & \multicolumn{8}{c|}{\textbf{MECO L2}} & \multicolumn{8}{c}{\textbf{OneStopL2}} \\
\multicolumn{1}{l}{} & \multicolumn{4}{c}{LexTALE} & \multicolumn{4}{c|}{Composite} & \multicolumn{4}{c}{LexTALE} & \multicolumn{4}{c}{Michigan} \\ \cmidrule{2-17}
\multicolumn{1}{l}{} & \multicolumn{2}{c}{Fixed} & \multicolumn{2}{c}{Any} & \multicolumn{2}{c}{Fixed} & \multicolumn{2}{c|}{Any} & \multicolumn{2}{c}{Fixed} & \multicolumn{2}{c}{Any} & \multicolumn{2}{c}{Fixed} & \multicolumn{2}{c}{Any} \\ \cmidrule{2-17}
\textbf{Features} & $r$ & MAE & $r$ & MAE & $r$ & MAE & $r$ & MAE & $r$ & MAE & $r$ & MAE & $r$ & MAE & $r$ & MAE \\
\midrule
Avg. Train & $0.00\phantom{^{***}}$ & $9.92\phantom{^{***}}$ & $0.00\phantom{^{***}}$ & $9.92\phantom{^{***}}$ & $0.00\phantom{^{***}}$ & $11.37\phantom{^{***}}$ & $0.00\phantom{^{***}}$ & $11.37\phantom{^{***}}$ & $-0.06\phantom{^{***}}$ & $10.32\phantom{^{**}}$ & $-0.02\phantom{^{*}}$ & $10.25\phantom{^{*}}$ & $-0.07\phantom{^{***}}$ & $7.52$ & $-0.01\phantom{^{*}}$ & $7.40$ \\
\meanbaselinerule
WPM & $0.48\phantom{^{***}}$ & $8.53\phantom{^{***}}$ & $0.48\phantom{^{***}}$ & $8.52\phantom{^{***}}$ & $0.44\phantom{^{***}}$ & $9.96\phantom{^{***}}$ & $0.44\phantom{^{***}}$ & $9.95\phantom{^{***}}$ & $0.42\phantom{^{***}}$ & $10.21\phantom{^{**}}$ & $0.47\phantom{^{*}}$ & $9.62\phantom{^{*}}$ & $0.37\phantom{^{***}}$ & $6.42$ & $0.47\phantom{^{*}}$ & $6.14$ \\
\midrule
Avg. Fix. & $0.55^{***}$ & $8.16^{***}$ & $0.54^{***}$ & $8.17^{***}$ & $0.54^{***}$ & $9.39^{***}$ & $\mathbf{0.54}^{***}$ & $9.41^{***}$ & $0.47\phantom{^{***}}$ & $10.00\phantom{^{**}}$ & $0.55^{*}$ & $9.04\phantom{^{*}}$ & $0.45^{*}\phantom{^{**}}$ & $6.34$ & $0.56^{*}$ & $5.87$ \\
S-Clusters & $0.56^{***}$ & $8.10^{***}$ & $0.55^{***}$ & $\mathbf{8.14}^{***}$ & $0.55^{***}$ & $9.26^{***}$ & $\mathbf{0.54}^{***}$ & $\mathbf{9.39}^{***}$ & $0.49\phantom{^{***}}$ & $9.75\phantom{^{**}}$ & $0.57^{*}$ & $8.67^{*}$ & $0.47^{*}\phantom{^{**}}$ & $6.26$ & $\mathbf{0.57}^{*}$ & $\mathbf{5.82}$ \\
WP-Coefs & $0.56^{***}$ & $8.12^{***}$ & $\mathbf{0.56}^{***}$ & $\mathbf{8.14}^{***}$ & $0.55^{***}$ & $9.30^{***}$ & $\mathbf{0.54}^{***}$ & $9.41^{***}$ & $0.48\phantom{^{***}}$ & $9.65\phantom{^{**}}$ & $\mathbf{0.58}^{*}$ & $\mathbf{8.57}^{*}$ & $0.46^{*}\phantom{^{**}}$ & $6.36$ & $0.56^{*}$ & $5.91$ \\
\midrule
Transitions & $0.55^{***}$ & $8.08^{***}$ & \phantom{0.}-\phantom{00} & \phantom{0.}-\phantom{00} & $0.56^{***}$ & $9.37^{***}$ & \phantom{0.}-\phantom{00} & \phantom{0.}-\phantom{00} & $0.50^{*}\phantom{^{**}}$ & $9.78\phantom{^{**}}$ & \phantom{0.}-\phantom{00} & \phantom{0.}-\phantom{00} & $0.47^{**}\phantom{^{*}}$ & $6.07$ & \phantom{0.}-\phantom{00} & \phantom{0.}-\phantom{00} \\
Word Fix. & $\mathbf{0.62}^{***}$ & $\mathbf{7.59}^{***}$ & \phantom{0.}-\phantom{00} & \phantom{0.}-\phantom{00} & $\mathbf{0.59}^{***}$ & $\mathbf{8.86}^{***}$ & \phantom{0.}-\phantom{00} & \phantom{0.}-\phantom{00} & $\mathbf{0.56}^{***}$ & $\mathbf{8.86}^{**}$ & \phantom{0.}-\phantom{00} & \phantom{0.}-\phantom{00} & $\mathbf{0.52}^{***}$ & $\mathbf{5.86}$ & \phantom{0.}-\phantom{00} & \phantom{0.}-\phantom{00} \\
\bottomrule
\end{tabular}
}
\caption{\textbf{Prediction of standard English proficiency test scores}. Pearson $r$ and Mean Absolute Error (MAE) for prediction of external proficiency scores using eye movements with Ridge regression. Presented are mean values using bootstrap. Statistically significant differences from the WPM baseline using a bootstrap test are marked with `$^{*}$' $p<0.05$, `$^{**}$' $p<0.01$, `$^{***}$' $p<0.001$. ``Fixed'' is the Fixed Text regime in which all the eye movement data is for the same texts presented to the test participant. ``Any'' is the Any Text regime in which no eye movement data is available for the texts of the test participant. The best result for each evaluation measure in each eye tracking dataset, reference proficiency test and Text regime is marked in bold.}
\label{tab:predictions_bootstrap}
\end{table*}

The main difference compared to \citet{berzak-etal-2018-assessing} is a much stronger performance of the WPM reading speed baseline ($r \leq 0.28$ in their data). We hypothesize that this difference could stem from strategic sentence rereading behavior for performing better on a question presented after each sentence in CELER, which is less feasible in passage reading in MECO L2 and OneStopL2. Despite this difference, eye movement feature sets systematically outperform WPM, confirming their value above and beyond reading speed. Similar results are obtained in the OneStopL2 information seeking regime (\Cref{tab:eyescore_seen_unseen_hunting_bootstrap} in \Cref{sec:app-info-seeking}), suggesting that EyeScore is robust to the reading manner.

\subsection{Predicting External Proficiency Scores}
\label{sec:prediction}

The results for predicting external proficiency test scores from eye movements of L2 participants are presented in \Cref{tab:predictions_bootstrap}. As can be expected from a model explicitly trained to predict external test outcomes, for any given Pearson $r$ evaluation, the correlations tend to be higher than those for EyeScore in \Cref{tab:eyescore_seen_unseen_bootstrap}, and the  improvements over the WPM baseline are more substantial. Here too, the results are in line with \citet{berzak-etal-2018-assessing}, with similar or higher correlations in most cases. The primary exception is OneStopL2 Fixed Text, where the results for both Michigan and LexTALE are low, even compared to the corresponding EyeScore results. We attribute this to the small training data in OneStopL2 for the Fixed Text regime (see \Cref{sec:eval-setting}), suggesting that differently from EyeScore, a sizable number of L2 training participants is needed for robust prediction of external scores. 

\Cref{tab:predictions-onestop-hunting} in \Cref{sec:app-info-seeking} presents similar results in information seeking, which suggest that the prediction of external test results largely generalizes to this reading regime. \Cref {sec:app-prediction-more-models} includes the results for LightGBM  (\Cref{tab:predictions_bootstrap_lgbm_holdout_meco,tab:predictions_bootstrap_lgbm_holdout_onestop}) and TabPFN-3 (\Cref{tab:predictions_bootstrap_tabpfn_meco,tab:predictions_bootstrap_tabpfn_onestop}). LightGBM yields similar performance to Ridge regression on MECO L2, and tends to perform worse on OneStopL2. TabPFN-3 largely outperforms Ridge regression on MECO L2, and performs similarly on OneStopL2.

\subsection{Comparison to Correlations between Standard Proficiency Tests}

Especially informative references for the presented results are the correlations between different standard proficiency tests. These can be thought of as an upper bound for EyeScore and as a target benchmark for the prediction task. In our datasets, the Pearson $r$ between LexTALE and the Composite score in MECO L2 is $0.55$, and  between LexTALE and Michigan in OneStopL2 is $0.61$. Reported correlations among widely used standardized English proficiency tests are typically in the range of $0.61$--$0.85$.\footnote{ EF SET has a correlation of $0.74$ with TOEFL iBT and $0.61$ with IELTS \citep{first2025ef}. The individual section correlations between IELTS and TOEFL iBT range from $0.68$ to $0.76$, with an overall test correlation of $0.85$ \citep{ikeda2025aligning}.}
The EyeScore correlations with standard tests in \Cref{tab:eyescore_seen_unseen_bootstrap} are lower, likely reflecting the radical difference in the testing methodology. Direct prediction of external test scores from eye movements is at the lower end of this range  for the best performing feature sets. Overall, these results strengthen the evidence for the viability of eye movement based language proficiency testing.

\section{Measuring and Mitigating L1 Bias in EyeScore}
\label{sec:bias}

Since EyeScore relies on similarity of reading patterns to the average English L1 reader, this measure might favor readers whose L1 is more similar to English. Below, we investigate this question in the Any Text regime.

\subsection{Measuring L1 Bias}
\label{sec:bias-measuring}

We use a two-step procedure for measuring EyeScore L1 bias. The procedure relies on a standard English proficiency test  which serves as a control for possible differences in English proficiency across the L1 groups in our sample of participants.
For a given external proficiency $\text{Test} \in \{\text{LexTALE},\text{Composite}, \text{Michigan}\}$, we fit an $\mathrm{EyeScore}_p \sim \mathrm{Test}_p$ regression on all L2 participants and obtain for each participant $p$ the residual by subtracting the model prediction from EyeScore:
\begin{equation}\label{eq:res}
\mathrm{res}_{p,\mathrm{Test}} \coloneqq \mathrm{EyeScore}_p - (\hat{\beta}_0 +  \hat{\beta}_{\mathrm{Test}}\cdot{\mathrm{Test}}_p)
\end{equation}
where $\hat{\beta}_0$ and $\hat{\beta}_{\mathrm{Test}}$ are the fitted regression coefficients and $\mathrm{Test}_p$ is the participant's test score. 

In the second step, we examine whether the residuals systematically depend on the linguistic proximity of the participant's L1s to English. To this end, we fit a regression model that predicts the residuals from the L1 distance from English $\mathrm{res}_{p,\mathrm{Test}} \sim d(\text{L1}_{p}, \text{English})$. The fitted model predicts a participant's residual based on their L1:
\begin{equation}\label{eq:res-pred}
\widehat{\mathrm{res}}_{p,\mathrm{Test}} = \hat{\beta}_0+\hat{\beta}_{dist}\cdot d(\text{L1}_p, \text{English})
\end{equation}
A slope coefficient $\hat{\beta}_{dist}$ which is significantly smaller than 0 would indicate that EyeScore is biased towards L1s that are closer to English, above and beyond the reference proficiency test.

We compute $d(\text{L1}, \text{English})$ as the angular distance between 
binarized language vectors with linguistic features from URIEL+  \citep[][v1.2]{khan-etal-2025-uriel}. We use 103 syntactic features from WALS \citep{matthew_dryer_2022_7385533}, Ethnologue \citep{lewis2014ethnologue}, and SSWL \citep{collins2009syntactic}, 28 phonological features from WALS and Ethnologue, and 42 writing system features from ScriptSource~\cite{Holloway2013}. 

\begin{figure}[ht!]
    \centering
    \includegraphics[width=1\linewidth]{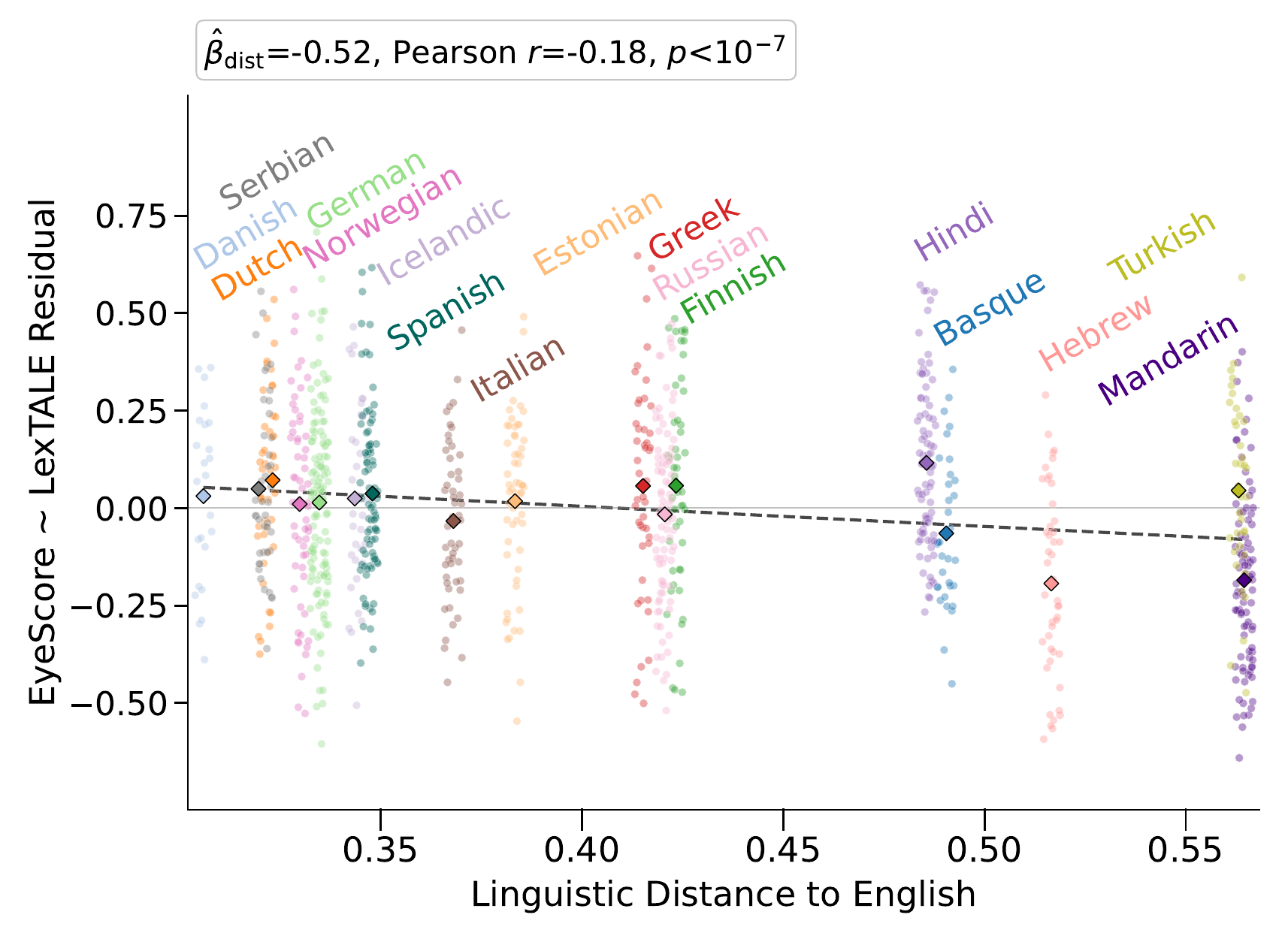}
    \caption{\textbf{L1-English Proximity Bias in EyeScore}. Residuals of an $\mathrm{EyeScore}_p \sim \mathrm{LexTALE}_p$ regression as a function of the linguistic distance of the participant's L1 to English: $\mathrm{res}_{p,\mathrm{LexTALE}} \sim d(\text{L1}_p, \text{English})$. The figure depicts MECO L2 participants in the Any Text regime using the WP-Coefs feature set. Each circle is a participant. Diamond shapes denote L1 means. Jitter is added for legibility. The dashed line is the model fit, which suggests that EyeScore favors participants whose L1 is more similar to English.}
    \label{fig:residual_regression}
\end{figure}

\begin{table*}[ht!]
\centering
\resizebox{\textwidth}{!}{%
\begin{tabular}{lccc|ccc|ccc|ccc}
\toprule
 & \multicolumn{6}{c|}{\textbf{MECO L2}} & \multicolumn{6}{c}{\textbf{OneStopL2}} \\
 & \multicolumn{3}{c}{LexTALE} & \multicolumn{3}{c|}{Composite} & \multicolumn{3}{c}{LexTALE} & \multicolumn{3}{c}{Michigan} \\ \cmidrule{2-13}
 & \multicolumn{1}{c}{EyeScore} & \multicolumn{2}{c|}{EyeScore\textsuperscript{db}} & \multicolumn{1}{c}{EyeScore} & \multicolumn{2}{c|}{EyeScore\textsuperscript{db}} & \multicolumn{1}{c}{EyeScore} & \multicolumn{2}{c|}{EyeScore\textsuperscript{db}} & \multicolumn{1}{c}{EyeScore} & \multicolumn{2}{c}{EyeScore\textsuperscript{db}} \\ \cmidrule{2-13}
\textbf{Features} &  & Seen L1 & New L1 &  & Seen L1 & New L1 &  & Seen L1 & New L1 &  & Seen L1 & New L1 \\
\midrule \cmidrule{1-13}
WPM & $-1.37^{***}$ & $-0.05$ & $-0.43$ & $-1.97^{***}$ & $-0.60\phantom{^{*}}$ & $-0.98^{**}$ & $-1.28$ & $\phantom{-}0.00$ & $-0.01$ & $-1.87^{*}$ & $-0.62$ & $-0.57$ \\
\midrule
Avg. Fix. & $-0.78^{***}$ & $-0.03$ & $-0.14$ & $-1.18^{***}$ & $-0.40\phantom{^{*}}$ & $-0.51^{*}\phantom{^{*}}$ & $-0.75$ & $\phantom{-}0.00$ & $\phantom{-}0.03$ & $-1.15\phantom{^{*}}$ & $-0.39$ & $-0.35$ \\
S-Clusters & $-0.43^{**}\phantom{^{*}}$ & $-0.02$ & $-0.11$ & $-0.71^{***}$ & $-0.28\phantom{^{*}}$ & $-0.37^{*}\phantom{^{*}}$ & $-0.53$ & $-0.01$ & $\phantom{-}0.03$ & $-0.88\phantom{^{*}}$ & $-0.36$ & $-0.30$ \\
WP-Coefs & $-0.52^{***}$ & $-0.02$ & $-0.10$ & $-0.71^{***}$ & $-0.19^{*}$ & $-0.27^{**}$ & $-0.63$ & $-0.01$ & $-0.01$ & $-0.88^{*}$ & $-0.25$ & $-0.24$ \\ \bottomrule
\end{tabular}%
}
\caption{\textbf{L1-English Proximity Bias in EyeScore and EyeScore\textsuperscript{db}} in the Any Text regime. Presented are $\hat{\beta}_{dist}$ coefficients from $\widehat{\mathrm{res}}_{p,\mathrm{Test}} = \hat{\beta}_0+\hat{\beta}_{dist}\cdot d(L1_p, \text{English})$ regressions, which capture EyeScore and EyeScore\textsuperscript{db} bias towards L1s that are linguistically more similar to English.
In all cases, EyeScore\textsuperscript{db} is obtained by debiasing EyeScore relative to LexTALE. Seen L1 / New L1 refers to whether the training data for deriving the debiasing transformation included participants from the same L1 as the test participant. Significance of a t-test 
for the $\hat{\beta}_{dist}$ coefficients is marked with  `$^{*}$'\,$p<0.05$, `$^{**}$'\,$p<0.01$, `$^{***}$'\,$p<0.001$. In MECO L2, the $\hat{\beta}_{dist}$ coefficients of EyeScore\textsuperscript{db} are significantly smaller than their EyeScore counterparts in all cases ($p<0.05$, bootstrap test).} 
\label{tab:lang-bias-triplet-coef-two_step}
\end{table*}

The $\mathrm{res}_{p,\text{LexTALE}} \sim d(\text{L1}_p, \text{English})$ regression fit for EyeScore using MECO L2 in the Any Text regime and WP-Coefs is presented in \Cref{fig:residual_regression}. We observe that participants whose L1 is less similar to English, indeed tend to score lower on EyeScore relative to the LexTALE reference ($\hat{\beta}_{dist}=-0.52, r=-0.18, p<10^{-7}$). \Cref{tab:lang-bias-triplet-coef-two_step} suggests that this negative correlation holds for all the feature sets relative to LexTALE and the Composite score in MECO L2, and relative to Michigan for WP-Coefs in OneStopL2. In the  other cases the coefficients are not significant, but consistently negative. \Cref{fig:residual_regression_meco,fig:residual_regression_onestop,fig:residual_regression_meco_composite,fig:residual_regression_onestop_michigan} in \Cref{sec:app-bias-obtain-methods-data} present the corresponding regression fits.

\subsection{Debiasing EyeScore}

To mitigate the L1 bias in EyeScore, we propose a simple correction procedure that subtracts from EyeScore the bias predicted by the participant's L1 distance to English. The debiased EyeScore is
\begin{equation}
\mathrm{EyeScore}^{\mathrm{db}}_p = \mathrm{EyeScore}_p -\widehat{\mathrm{res}}_{p,\mathrm{Test}}
\end{equation}
where $\widehat{\mathrm{res}}_{p,\mathrm{Test}}$ is the predicted mean bias for the participant's L1 relative to an external test.

EyeScore\textsuperscript{db} is computed for test set participants by adjusting their EyeScore using a $\widehat{\mathrm{res}}_{p,\mathrm{Test}}$ model fitted on a training set using $k$-fold cross validation, where $k$ is the number of L1s. We consider two data split variants, \textit{Seen L1}: stratified k-fold cross validation with all the L1s in the training set and test set of each split, and a more stringent \textit{New L1} split: $k$-1 L1s in the training set and a held out L1 in the test set. 

To evaluate the debiasing procedure, we refit the models from \Cref{eq:res,eq:res-pred} using EyeScore\textsuperscript{db}, and compare the resulting $\hat{\beta}_{dist}$ coefficients to those fitted with EyeScore. In our primary evaluation,  EyeScore\textsuperscript{db} is obtained by debiasing EyeScore relative to LexTALE (i.e.  $\mathrm{EyeScore}-\widehat{\mathrm{res}}_{p,\mathrm{LexTALE}}$). Thus, for LexTALE, the same test is used for debiasing and for evaluation, while the Composite and Michigan evaluations capture cross-test robustness. 

\Cref{tab:lang-bias-triplet-coef-two_step} presents the resulting $\hat{\beta}_{dist}$ coefficients in the Any Text regime. \Cref{sec:app-bias-bias} \Cref{tab:lang-bias-any-pearson} presents the corresponding Pearson $r$ of the models. For LexTALE, we find that the bias is eliminated nearly entirely on both MECO L2 and OneStopL2, especially in the Seen L1 evaluations. Most of the bias is also removed when the LexTALE-debiased scores are evaluated against the other two external tests, Michigan and Composite. This outcome suggests cross-test robustness in debiasing and evaluation. Here too, the Seen L1 evaluations are more effective compared to New L1. A similar pattern of results is obtained  when EyeScore\textsuperscript{db} is obtained by debiasing EyeScore relative to Composite (\Cref{sec:app-bias-obtain-by} \Cref{tab:lang-bias-triplet-coef-two_step-two_step-compositedb-full-any-coef}) and relative to Michigan (\Cref{sec:app-bias-obtain-by} \Cref{tab:lang-bias-triplet-coef-two_step-two_step-michigandb-full-any-coef}). 

\Cref{sec:app-bias-additional-results} presents largely similar results for the Fixed Text regime (\Cref{tab:lang-bias-fixed-coef}) and for information seeking reading in OneStopL2 (\Cref{tab:lang-bias-triplet-coef-two_step-hunting}). Overall, the results indicate that the debiasing procedure is effective in mitigating L1 bias. Finally, we note that a similar procedure for L1 bias diagnosis and removal can be implemented with a single multiple regression model of the form \mbox{$ \mathrm{EyeScore}_p \sim \mathrm{Test}_{p} + d(\text{L1}_p, \text{English})$}.
 
\subsection{Does Debiasing EyeScore Hurt Correlations with Standard Tests?} 
\label{sec:debiasing-validity}

This is something you might be wondering about. To examine this question, in \Cref{sec:bias-hurt-corr} \Cref{tab:eyescore_seen_unseen-debiased-lextale-diff} we present $\Delta r_{\mathrm{db}}$, the difference for L2 participants between (1) the Pearson $r$ of an EyeScore\textsuperscript{db} debiased relative to LexTALE with each of the standard proficiency tests, and (2) the corresponding Pearson $r$ for EyeScore (\Cref{tab:eyescore_seen_unseen_bootstrap}). Although significant in some cases, the absolute differences in Pearson $r$ between the two measures are small. In the Any Text regime, the average $\Delta r_{\mathrm{db}}$ across feature sets for the correlation with LexTALE is $-0.008$ for MECO L2 and $-0.016$ for OneStopL2. The average $\Delta r_{\mathrm{db}}$ with the Composite score is $+0.006$, and $-0.007$ with Michigan. Similar results are obtained in the Fixed Text regime. We further note that the average correlation between EyeScore and EyeScore\textsuperscript{db} across all feature sets in both datasets is $0.99$, indicating the overall stability of the debiasing procedure. 

\section{EyeScore Reliability}
\label{sec:reliability}

An important psychometric indicator for the quality of a test is its reliability, namely the extent to which participant scores are consistent across different variants of the test. Despite the significance of this test characteristic, the reliability of eye-tracking-based proficiency tests has yet to be measured. 
Here, we examine two standard measures of reliability: internal consistency and split-half reliability, with respect to EyeScore. 

\subsection{Internal Consistency} 
\label{sec:internal-consistency}

Internal consistency measures the degree to which items in a test capture the same underlying construct. We use Cronbach's $\alpha$ \citep{cronbach1951coefficient}, a standard measure of internal consistency, defined as     \mbox{$\alpha = \frac{n}{n-1}(1-\frac{\sum_{i=1}^n{V_i}}{V_t})$}, where $n$ is the number of test items, $V_i$ is the variance of test item $i$ scores across participants and $V_t$ is the variance of the test scores across participants. The range of possible values for this measure is 0 to 1. We also report the standard error of measurement (SEM) \citep{harvill1991standard} for Cronbach's $\alpha$, which is used to estimate measure uncertainty, and defined as $\mathrm{SD}\cdot\sqrt{1-\alpha}$, where SD is the standard deviation estimate of test scores across participants.

Cronbach's $\alpha$ requires the test to consist of multiple items, a score for each participant on each item, and an overall test score for each participant based on an aggregation of the item-level scores. 
A possible unit for an item that applies to both MECO L2 and OneStopL2 is eye movements over a \emph{passage}. Accordingly, we construct EyeScore$^\mathrm{items}$, an aggregated-per-item variant of EyeScore, where we first compute for each participant an EyeScore for each passage, and then define the overall EyeScore of the participant as the mean of the item-level scores. EyeScore$^\mathrm{items}$ correlates very strongly with EyeScore (average Pearson $r$ 0.983, \Cref{sec:eyescoreitems-eyescore-comparison} \Cref{tab:original_vs_per_item_eyescore}) and has the same or slightly higher correlations with external proficiency tests (average $\Delta r$ +0.011, \Cref{sec:eyescoreitems-eyescore-comparison} \Cref{tab:evaluation-eyescore-delta}). Since Cronbach's $\alpha$ assumes that all participants have scores for all the items, in MECO L2 we use only the 144 participants who have eye movement data for all 12 passages. Additional details on the evaluation of internal consistency are provided in \Cref{sec:app-eval-setting-per-item}.

\subsection{Split-Half Reliability} 
This measure is a proxy for test-retest reliability, where instead of taking the test twice, a single test is divided into two halves. Then, the Pearson $r$ correlation between the test scores on each half is computed across participants.
As the results of a split-half analysis can vary across splits \citep{cronbach1951coefficient}, we report the mean over the Pearson $r$ coefficients of 20 random splits of the passages. Additional details on the split-half reliability measurement procedure are provided in \Cref{sec:app-eval-setting-spltihalf}.

\subsection{Results}

The Cronbach's $\alpha$ and split-half reliability of EyeScore in the Any Text regime are presented in \Cref{tab:eyescore-reliability-any-paragraph}. We find that in nearly all cases, the reliability of EyeScore is extremely high according to both measures, with an average Cronbach's $\alpha$ of $0.98$ and an average Pearson $r$ of split-half reliability of $0.93$. Reliability analyses in the Fixed Text regime (\Cref{sec:app-reliability-eyescore-extended} \Cref{tab:eyescore-reliability-fixed-paragraph}) and for the information seeking reading regime in OneStopL2 (\Cref{sec:app-reliability-eyescore-extended} \Cref{tab:eyescore-reliability-regimes-paragraph-information-seeking}) yield similar results. 

\begin{table}[ht!]
\centering
\resizebox{\columnwidth}{!}{
\begin{tabular}{lccc|ccc}
\toprule
 \multicolumn{1}{c}{} & \multicolumn{3}{c|}{\textbf{MECO L2}} & \multicolumn{3}{c}{\textbf{OneStopL2}} \\
\textbf{Features} & Cronbach's $\alpha$ & SEM & Split-Half $r$ & Cronbach's $\alpha$ & SEM & Split-Half $r$ \\
\midrule
WPM & 0.98 & 0.13 & 0.93 & 0.99 & 0.07 & 0.98 \\
\midrule
Avg. Fix. & 0.98 & 0.08 & 0.92 & 0.99 & 0.04 & 0.98 \\
S-Clusters & 0.98 & 0.04 & 0.93 & 1.00 & 0.02 & 0.98 \\
WP-Coefs & 0.91 & 0.05 & 0.79 & 0.99 & 0.02 & 0.95 \\
\bottomrule
\end{tabular}
}
\caption{\textbf{Reliability of EyeScore} for L2 participants in the Any Text regime.}
\label{tab:eyescore-reliability-any-paragraph}
\end{table}

Importantly, the reliability of EyeScore tends to be \emph{higher} than the reliability of standardized English proficiency tests. In  OneStopL2, the Michigan test has a Cronbach's $\alpha$ of $0.91$ (stratified), and a split-half Pearson $r$ of $0.92$. The average Cronbach's $\alpha$ across the IELTS reading and listening sections is $0.91$ \citep{IELTS2026}. The overall reliability of TOEFL-iBT is $0.90$ \cite{manna2025toefl}. Additional details on the reliability of these tests are provided in \Cref{sec:app-reliability-comparison-other}.

\section{Conclusion}

We present the first replication study of a language proficiency testing paradigm in which instead of standard test items, proficiency is derived from eye movements in reading. We find that this methodology is effective in passage reading across datasets, standard proficiency references and reading regimes. We further investigate validity with respect to L1 bias and the reliability of the method.  

Overall, our experiments provide substantial support for the viability of eye movement based proficiency testing. In the future, we envision that it can be deployed either as a standalone technology or as an add-on to traditional proficiency tests. The limitations section below details some of the work that is needed in order to make this paradigm viable for real-world settings.

\section{Limitations}

Our study has a number of limitations. Perhaps the cardinal limitation of eye tracking based approaches to language proficiency assessment is that they can be used for assessing only \emph{comprehension} related linguistic abilities. Production abilities for spoken and written language currently remain out of scope for such methods. 

Our L1 bias correction procedure has several limitations. First, it relies on the estimation of L1 distances from English, which can vary according to the chosen set of language features, how missing feature values are handled, and which metric is used to measure language distances. Another limitation is the dependence on an external reference test, whose choice can also influence the procedure outcomes. Even more importantly, we rely on the assumption that the external tests are not themselves biased towards specific L1s, and in particular towards L1s that are closer to English. If this assumption does not hold, our procedure will underestimate the amount of L1 bias that needs to be removed. Finally, our debiasing pipeline assumes that participants have a single L1, and thus cannot be straightforwardly applied to participants with several L1s.

Additional limitations concern the available data. While the recent introduction of MECO L2, OneStopL2 and OneStop has considerably expanded the possibilities for developing eye movement based language proficiency testing, these datasets are restricted to English, cover only two textual domains, and include participants from specific language backgrounds and a restricted range of ages. A more general investigation would require the collection and analysis of data for additional L2 target languages and textual domains, additional participant populations such as children, and participants from additional native language backgrounds.

It is further important to note that the used datasets were collected with state-of-the-art eye tracking equipment in laboratory conditions. To establish performance in real-world scenarios, additional work is needed with lower-quality eyetracking equipment and in settings outside of the lab. Finally, further data collection and experimentation are needed to characterize additional properties of eye movement based language proficiency tests, such as the extent to which participants can cheat in such tests, for example, by using top-down reading strategies such as skimming the text, as well as test reliability across different test sessions.

\section{Ethical Considerations}

All three eyetracking datasets used in this study, MECO L2, OneStopL2 and OneStop, were collected under institutional IRB protocols. All the participants have provided written consent for participation. All the data is anonymized. Studying the relation between eye movements and linguistic proficiency is among the intended use cases for which the datasets were collected. We further note that while it has been demonstrated that eyetracking data can be used for reader identification \cite[e.g.][]{jager2019deep}, the anonymization of the used datasets precludes such uses.

\section*{Acknowledgments}
This work was supported by ISF grant 1499/22.

\bibliography{references}

@article{bllip2000,
  title={{BLLIP} 1987-89 {WSJ} corpus release 1},
  author={Charniak, Eugene and Blaheta, Don and Ge, Niyu and Hall, Keith and Hale, John and Johnson, Mark},
  journal={Linguistic Data Consortium, Philadelphia},
  volume={36},
  year={2000}
}

@article{jeon2014l2,
  title={L2 reading comprehension and its correlates: A meta-analysis},
  author={Jeon, Eun Hee and Yamashita, Junko},
  journal={Language learning},
  volume={64},
  number={1},
  pages={160--212},
  year={2014},
  publisher={Wiley Online Library}
}

@inproceedings{ke2017lightgbm,
 author = {Ke, Guolin and Meng, Qi and Finley, Thomas and Wang, Taifeng and Chen, Wei and Ma, Weidong and Ye, Qiwei and Liu, Tie-Yan},
 booktitle = {Advances in Neural Information Processing Systems},
 editor = {I. Guyon and U. Von Luxburg and S. Bengio and H. Wallach and R. Fergus and S. Vishwanathan and R. Garnett},
 pages = {},
 publisher = {Curran Associates, Inc.},
 title = {{LightGBM}: A Highly Efficient Gradient Boosting Decision Tree},
 url = {https://proceedings.neurips.cc/paper_files/paper/2017/file/6449f44a102fde848669bdd9eb6b76fa-Paper.pdf},
 volume = {30},
 year = {2017}
}

@misc{grinsztajn2026tabpfn3technicalreport,
      title={{TabPFN}-3: Technical Report}, 
      author={Léo Grinsztajn and Klemens Flöge and Oscar Key and Felix Birkel and Philipp Jund and Brendan Roof and Mihir Manium and Shi Bin Hoo and Magnus Bühler and Anurag Garg and Dominik Safaric and Jake Robertson and Benjamin Jäger and Simone Alessi and Adrian Hayler and Vladyslav Moroshan and Lennart Purucker and Philipp Singer and Alan Arazi and Julien Siems and Jan Hendrik Metzen and Georg Grab and Nick Erickson and Siyuan Guo and Eliott Kalfon and Simon Bing and David Salinas and Clara Cornu and Lilly Charlotte Wehrhahn and Diana Kriuchkova and Kursat Kaya and Lydia Sidhoum and Marie Salmon and Jerry Chen and Madelon Hulsebos and Yann LeCun and Samuel Müller and Bernhard Schölkopf and Sauraj Gambhir and Noah Hollmann and Frank Hutter},
      year={2026},
      eprint={2605.13986},
      archivePrefix={arXiv},
      primaryClass={cs.LG},
      url={https://arxiv.org/abs/2605.13986}, 
}

@manual{Holloway2013,
  author       = {Holloway, Steph},
  title        = {{ScriptSource} -- Writing systems, computers and people},
  organization = {SIL International},
  year         = {2013},
  url          = {https://scriptsource.org/cms/scripts/page.php},
}

@misc{yoav_meiri_2026_20376826,
  author       = {Yoav Meiri and Omer Shubi and Keren Gruteke Klein and
                  David R. Reich and Yevgeni Berzak},
  title        = {{lacclab/psycholing-metrics: v1.1.9}},
  howpublished = {Zenodo},
  year         = {2026},
  month        = may,
  note         = {\url{https://doi.org/10.5281/zenodo.20376826}},
}

@misc{collins2009syntactic,
  title = {Syntactic Structures of the World's Languages ({SSWL})},
  author={Collins, Chris and Kayne, Richard},
  year={2009},

}

@misc{lewis2014ethnologue,
  title     = {Ethnologue: Languages of the World},
  author    = {Lewis, M. Paul and Simons, Gary F. and Fennig, Charles D.},
  year      = {2015},
  url={http://www.ethnologue.com}
}

@inproceedings{khan-etal-2025-uriel,
    title = "{URIEL}+: Enhancing Linguistic Inclusion and Usability in a Typological and Multilingual Knowledge Base",
    author = {Khan, Aditya  and
      Shipton, Mason  and
      Anugraha, David  and
      Duan, Kaiyao  and
      Hoang, Phuong H.  and
      Khiu, Eric  and
      Do{\u{g}}ru{\"o}z, A. Seza  and
      Lee, En-Shiun Annie},
    booktitle = "Proceedings of the 31st International Conference on Computational Linguistics",
    month = jan,
    year = "2025",
    publisher = "Association for Computational Linguistics",
    url = "https://aclanthology.org/2025.coling-main.463/",
    pages = "6937--6952",
}

@misc{matthew_dryer_2022_7385533,
  author       = {Matthew Dryer and
                  Martin Haspelmath},
  title        = {The World Atlas of Language Structures Online},
  month        = dec,
  year         = 2022,
  publisher    = {Zenodo},
  version      = {v2020.3},
  doi          = {10.5281/zenodo.7385533},
  url          = {https://doi.org/10.5281/zenodo.7385533},
}

@article{CLAHSEN_FELSER_2006, 
title={Grammatical processing in language learners}, 
volume={27}, 
DOI={10.1017/S0142716406060024}, 
number={1}, 
journal={Applied Psycholinguistics}, 
author={Clahsen, Harald and Felser, Claudia}, 
year={2006}, 
pages={3–42}}

@inproceedings{berzak-etal-2018-assessing,
    title = "Assessing Language Proficiency from Eye Movements in Reading",
    author = "Berzak, Yevgeni  and
      Katz, Boris  and
      Levy, Roger",
    booktitle = "Proceedings of the 2018 Conference of the North {A}merican Chapter of the Association for Computational Linguistics: Human Language Technologies, Volume 1 (Long Papers)",
    month = jun,
    year = "2018",
    address = "New Orleans, Louisiana",
    publisher = "Association for Computational Linguistics",
    url = "https://aclanthology.org/N18-1180/",
    doi = "10.18653/v1/N18-1180",
    pages = "1986--1996",
 
}

@inproceedings{reich2022inferring,
  title={Inferring native and non-native human reading comprehension and subjective text difficulty from scanpaths in reading},
  author={Reich, David Robert and Prasse, Paul and Tschirner, Chiara and Haller, Patrick and Goldhammer, Frank and J{\"a}ger, Lena A},
  booktitle={2022 Symposium on eye tracking research and applications},
  pages={1--8},
  year={2022}
}

@article{meziere2023using,
  author = {Mézière, Diane C. and Lili Yu and Reichle, Erik D. and {von der Malsburg}, Titus and Genevieve McArthur},
  title = {Using eye-tracking measures to predict reading comprehension},
  journal = {Reading Research Quarterly},
  year = {2023},
  volume = {58},
  number = {3},
  pages = {425-449},
  doi = {10.1002/rrq.498},
  url = {https://ila.onlinelibrary.wiley.com/doi/abs/10.1002/rrq.498}               ,
}

@article{copeland2014,
  title={Predicting reading comprehension scores from eye movements using artificial neural networks and fuzzy output error.},
  author={Copeland, Leana and Gedeon, Tom and Mendis, Sumudu},
  journal={Artificial Intelligence Research},
  volume={3},
  number={3},
  pages={35--48},
  year={2014}
}

@InProceedings{Ahn2020TowardsBehavior,
  author    = {Ahn, Seoyoung and Kelton, Conor and Balasubramanian, Aruna and Zelinsky, Greg},
  title     = {Towards predicting reading comprehension from gaze behavior},
  booktitle = {Symposium on Eye Tracking Research and Applications},
  year      = {2020},
  pages     = {1--5},
  month     = feb,
  publisher = {Association for Computing Machinery},
}

@inproceedings{shubi2024fine,
  title={Fine-grained prediction of reading comprehension from eye movements},
  author={Shubi, Omer and Meiri, Yoav and Hadar, Cfir Avraham and Berzak, Yevgeni},
  booktitle={Proceedings of the 2024 Conference on Empirical Methods in Natural Language Processing},
  pages={3372--3391},
  year={2024}
}

@inproceedings{skerath2023native,
  title={Native language prediction from gaze: a reproducibility study},
  author={Skerath, Lina and Toborek, Paulina and Zieli{\'n}ska, Anita and Barrett, Maria and Van Der Goot, Rob},
  booktitle={Proceedings of the 61st Annual Meeting of the Association for Computational Linguistics (Volume 4: Student Research Workshop)},
  pages={152--159},
  year={2023}
}

@inproceedings{berzak2017predicting,
  title={Predicting native language from gaze},
  author={Berzak, Yevgeni and Nakamura, Chie and Flynn, Suzanne and Katz, Boris},
  booktitle={Proceedings of the 55th Annual Meeting of the Association for Computational Linguistics (Volume 1: Long Papers)},
  pages={541--551},
  year={2017}
}

@inproceedings{shubi2025decoding,
  title={Decoding reading goals from eye movements},
  author={Shubi, Omer and Hadar, Cfir Avraham and Berzak, Yevgeni},
  booktitle={Proceedings of the 63rd Annual Meeting of the Association for Computational Linguistics (Volume 1: Long Papers)},
  pages={5616--5637},
  year={2025}
}

@inproceedings{shubi2025eyebench,
  title={{EyeBench}: Predictive Modeling from Eye Movements in Reading},
  author={Shubi, Omer and Reich, David Robert and Klein, Keren Gruteke and Angel, Yuval and Prasse, Paul and J{\"a}ger, Lena Ann and Berzak, Yevgeni},
  year={2025},
  booktitle={The Thirty-ninth Annual Conference on Neural Information Processing Systems Datasets and Benchmarks Track}
}

@inproceedings{rello2015detecting,
  title={Detecting readers with dyslexia using machine learning with eye tracking measures},
  author={Rello, Luz and Ballesteros, Miguel},
  booktitle={Proceedings of the 12th international web for all conference},
  pages={1--8},
  year={2015}
}

@inproceedings{haller2022dyslexia,
  title={Eye-tracking based classification of {Mandarin} {Chinese} readers with and without dyslexia using neural sequence models},
  author={Haller, Patrick and S{\"a}uberli, Andreas and Kiener, Sarah and Pan, Jinger and Yan, Ming and J{\"a}ger, Lena},
  booktitle={Proceedings of the Workshop on Text Simplification, Accessibility, and Readability (TSAR-2022)},
  pages={111--118},
  year={2022}
}

@inproceedings{laurinavichyute2025automatic,
  title={Automatic detection of dyslexia based on eye movements during reading in {Russian}},
  author={Laurinavichyute, Anna and Lopukhina, Anastasiya and Reich, David Robert},
  booktitle={Proceedings of the 63rd Annual Meeting of the Association for Computational Linguistics (Volume 2: Short Papers)},
  pages={59--66},
  year={2025}
}

@inproceedings{meiri2025deja,
  title={D{\'e}j{\`a} Vu? {Decoding} Repeated Reading from Eye Movements},
  author={Meiri, Yoav and Shubi, Omer and Hadar, Cfir Avraham and Nitzav, Ariel Kreisberg and Berzak, Yevgeni},
  booktitle={Proceedings of the 63rd Annual Meeting of the Association for Computational Linguistics (Volume 1: Long Papers)},
  pages={19460--19482},
  year={2025}
}

@article{just1980theory,
  title={A theory of reading: from eye fixations to comprehension.},
  author={Just, Marcel A and Carpenter, Patricia A},
  journal={Psychological review},
  volume={87},
  number={4},
  pages={329},
  year={1980},
  publisher={American Psychological Association}
}

@article{reichle1998toward,
  title={Toward a model of eye movement control in reading.},
  author={Reichle, Erik D and Pollatsek, Alexander and Fisher, Donald L and Rayner, Keith},
  journal={Psychological review},
  volume={105},
  number={1},
  pages={125},
  year={1998},
  publisher={American Psychological Association}
}

@book{rayner2012psychology,
  title={Psychology of reading},
  author={Rayner, Keith and Pollatsek, Alexander and Ashby, Jane and Clifton Jr, Charles},
  year={2012},
  publisher={Psychology Press}
}

@article{engbert2005swift,
  title={{SWIFT}: a dynamical model of saccade generation during reading.},
  author={Engbert, Ralf and Nuthmann, Antje and Richter, Eike M and Kliegl, Reinhold},
  journal={Psychological review},
  volume={112},
  number={4},
  pages={777},
  year={2005},
  publisher={American Psychological Association}
}

@article{demberg2008data,
  title={Data from eye-tracking corpora as evidence for theories of syntactic processing complexity},
  author={Demberg, Vera and Keller, Frank},
  journal={Cognition},
  volume={109},
  number={2},
  pages={193--210},
  year={2008},
  publisher={Elsevier}
}

@article{onestop2025,
author={Berzak, Yevgeni and Malmaud, Jonathan and Shubi, Omer and Meiri, Yoav and Lion, Ella and Levy, Roger},
title={{OneStop}: A 360-Participant {English} Eye Tracking Dataset with Different Reading Regimes},
journal={Scientific Data},
year={2025},
doi={10.1038/s41597-025-06272-2},
}

@article{mecoL2-2023,
  title={Text reading in {English} as a second language: Evidence from the Multilingual Eye-Movements Corpus},
  author={Kuperman, Victor and Siegelman, Noam and Schroeder, Sascha and Acart{\"u}rk, Cengiz and Alexeeva, Svetlana and Amenta, Simona and Bertram, Raymond and Bonandrini, Rolando and Brysbaert, Marc and Chernova, Daria and others},
  journal={Studies in second language acquisition},
  volume={45},
  number={1},
  pages={3--37},
  year={2023},
  publisher={Cambridge University Press}
}

@article{mecoL2-2025,
  title={New data on text reading in {English} as a second language: The Wave 2 expansion of the Multilingual Eye-Movement Corpus ({MECO})},
  author={Kuperman, Victor and Schroeder, Sascha and Acart{\"u}rk, Cengiz and Agrawal, Niket and Alexandre, Dominick M and Bolliger, Lena S and Brasser, Jan and Campos-Rojas, C{\'e}sar and Drieghe, Denis and {\DJ}ur{\dj}evi{\'c}, Du{\v{s}}ica Filipovi{\'c} and others},
  journal={Studies in Second Language Acquisition},
  volume={47},
  number={2},
  pages={677--695},
  year={2025},
  publisher={Cambridge University Press}
}

@article{lemhofer2012introducing,
  title={Introducing {LexTALE}: A quick and valid lexical test for advanced learners of {English}},
  author={Lemh{\"o}fer, Kristin and Broersma, Mirjam},
  journal={Behavior research methods},
  volume={44},
  number={2},
  pages={325--343},
  year={2012},
  publisher={Springer}
}

@article{kliegl2004length,
  title={Length, frequency, and predictability effects of words on eye movements in reading},
  author={Kliegl, Reinhold and Grabner, Ellen and Rolfs, Martin and Engbert, Ralf},
  journal={European journal of cognitive psychology},
  volume={16},
  number={1-2},
  pages={262--284},
  year={2004},
  publisher={Taylor \& Francis}
}

@article{rayner2004effects,
  title={The effects of frequency and predictability on eye fixations in reading: implications for the {E-Z} {Reader} model.},
  author={Rayner, Keith and Ashby, Jane and Pollatsek, Alexander and Reichle, Erik D},
  journal={Journal of Experimental Psychology: Human Perception and Performance},
  volume={30},
  number={4},
  pages={720},
  year={2004},
  publisher={American Psychological Association}
}

@article{santorini1990part,
  title={Part-of-speech tagging guidelines for the {Penn} {Treebank} project (3rd revision, 2nd printing)},
  author={Santorini, Beatrice},
  journal={Ms., Department of Linguistics, UPenn. Philadelphia, PA},
  year={1990}
}

@inproceedings{petrov2012universal,
    title = "A Universal Part-of-Speech Tagset",
    author = "Petrov, Slav  and
      Das, Dipanjan  and
      McDonald, Ryan",
    booktitle = "Proceedings of the Eighth International Conference on Language Resources and Evaluation ({LREC}'12)",
    month = may,
    year = "2012",
    address = "Istanbul, Turkey",
    publisher = "European Language Resources Association (ELRA)",
    url = "https://aclanthology.org/L12-1115/",
    pages = "2089--2096",
}

@misc{robyn_speer_2018,
  author       = {Robyn Speer and
                  Joshua Chin and
                  Andrew Lin and
                  Sara Jewett and
                  Lance Nathan},
  title        = {{LuminosoInsight}/wordfreq: v2.2},
  month        = oct,
  year         = 2018,
  doi          = {10.5281/zenodo.1443582},
  url          = {https://doi.org/10.5281/zenodo.1443582}
}

@article{cronbach1951coefficient,
  title={Coefficient alpha and the internal structure of tests},
  author={Cronbach, Lee J},
  journal={Psychometrika},
  volume={16},
  number={3},
  pages={297--334},
  year={1951},
  publisher={Springer-Verlag}
}

@article{berzak-levy2023,
    author = {Berzak, Yevgeni and Levy, Roger},
    title = {Eye Movement Traces of Linguistic Knowledge in Native and Non-Native Reading},
    journal = {Open Mind},
    volume = {7},
    pages = {179-196},
    year = {2023},
    month = {06},
    issn = {2470-2986},
    doi = {10.1162/opmi_a_00084},
    url = {https://doi.org/10.1162/opmi_a_00084},
    eprint = {https://direct.mit.edu/opmi/article-pdf/doi/10.1162/opmi_a_00084/2133839/opmi_a_00084.pdf},
}

@article{cop2015frequency,
  title={Frequency effects in monolingual and bilingual natural reading},
  author={Cop, Uschi and Keuleers, Emmanuel and Drieghe, Denis and Duyck, Wouter},
  journal={Psychonomic bulletin \& review},
  volume={22},
  number={5},
  pages={1216--1234},
  year={2015},
  publisher={Springer}
}

@article{whitford2012second,
  title={Second-language experience modulates first-and second-language word frequency effects: Evidence from eye movement measures of natural paragraph reading},
  author={Whitford, Veronica and Titone, Debra},
  journal={Psychonomic bulletin \& review},
  volume={19},
  number={1},
  pages={73--80},
  year={2012},
  publisher={Springer}
}

@article{whitford2017effects,
  title={The effects of word frequency and word predictability during first-and second-language paragraph reading in bilingual older and younger adults.},
  author={Whitford, Veronica and Titone, Debra},
  journal={Psychology and aging},
  volume={32},
  number={2},
  pages={158},
  year={2017},
  publisher={American Psychological Association}
}

@article{cop2015eye,
  title={Eye movement patterns in natural reading: A comparison of monolingual and bilingual reading of a novel},
  author={Cop, Uschi and Drieghe, Denis and Duyck, Wouter},
  journal={PloS one},
  volume={10},
  number={8},
  pages={e0134008},
  year={2015},
  publisher={Public Library of Science San Francisco, CA USA}
}

@article{berzak2022celer,
  title={{CELER}: A 365-participant corpus of eye movements in {L1} and {L2} {English} reading},
  author={Berzak, Yevgeni and Nakamura, Chie and Smith, Amelia and Weng, Emily and Katz, Boris and Flynn, Suzanne and Levy, Roger},
  journal={Open Mind},
  volume={6},
  pages={41--50},
  year={2022},
  publisher={MIT Press One Broadway, 12th Floor, Cambridge, Massachusetts 02142, USA~…}
}

@article{radford_language_2019,
  title={Language models are unsupervised multitask learners},
  author={Radford, Alec and Wu, Jeffrey and Child, Rewon and Luan, David and Amodei, Dario and Sutskever, Ilya},
  journal={OpenAI blog},
  volume={1},
  number={8},
  pages={9},
  year={2019}
}

@article{schotter2025beginner,
  title={A beginner’s guide to eye tracking for psycholinguistic studies of reading},
  author={Schotter, Elizabeth R and Dillon, Brian},
  journal={Behavior research methods},
  volume={57},
  number={2},
  pages={68},
  year={2025},
  publisher={Springer}
}

@article{chuang2025,
title = {Language assessment in the era of generative artificial intelligence: Opportunities, challenges, and future directions},
journal = {System},
volume = {134},
pages = {103846},
year = {2025},
issn = {0346-251X},
doi = {https://doi.org/10.1016/j.system.2025.103846},
url = {https://www.sciencedirect.com/science/article/pii/S0346251X25002568},
author = {Ping-Lin Chuang and Xun Yan},
}

@article{odlin1989language,
  title={Language transfer},
  author={Odlin, Terence},
  journal={Cambridge, UK: Cambridge},
  year={1989}
}

@book{jarvis2008crosslinguistic,
  title={Crosslinguistic influence in language and cognition},
  author={Jarvis, Scott and Pavlenko, Aneta},
  year={2008},
  publisher={Routledge}
}

@misc{honnibal2020spacy,
author = {Honnibal, Matthew and Montani, Ines and Van Landeghem, Sofie and Boyd, Adriane},
doi = {10.5281/zenodo.1212303},
title = {{spaCy: Industrial-strength Natural Language Processing in Python}},
year = {2020}
}

@inproceedings{jager2019deep,
  title={Deep eyedentification: Biometric identification using micro-movements of the eye},
  author={J{\"a}ger, Lena A and Makowski, Silvia and Prasse, Paul and Liehr, Sascha and Seidler, Maximilian and Scheffer, Tobias},
  booktitle={Joint European Conference on Machine Learning and Knowledge Discovery in Databases},
  pages={299--314},
  year={2019},
  organization={Springer}
}

@article{andrews2010lexical,
  title={Lexical precision in skilled readers: Individual differences in masked neighbor priming},
  author={Andrews, Sally and Hersch, Jolyn},
  journal={Journal of Experimental Psychology: General},
  volume={139},
  number={2},
  pages={299},
  year={2010},
  publisher={American Psychological Association}
}

@article{beglar2010rasch,
  title={{A Rasch-based validation of the Vocabulary Size Test}},
  author={Beglar, David},
  journal={Language testing},
  volume={27},
  number={1},
  pages={101--118},
  year={2010},
  publisher={SAGE Publications Sage UK: London, England}
}

@book{torgesen1999towre,
  title={TOWRE: Test of word reading efficiency},
  author={Torgesen, Joseph K and Rashotte, Carol Alexander and Wagner, Richard K},
  year={1999},
  publisher={Pro-ed Austin, TX}
}

@article{harvill1991standard,
  title={Standard error of measurement: an {NCME} instructional module on},
  author={Harvill, Leo M},
  journal={Educational Measurement: issues and practice},
  volume={10},
  number={2},
  pages={33--41},
  year={1991},
  publisher={Wiley Online Library}
}

@article{kolen1996conditional,
  title={Conditional standard errors of measurement for scale scores using {IRT}},
  author={Kolen, Michael J and Zeng, Lingjia and Hanson, Bradley A},
  journal={Journal of Educational Measurement},
  volume={33},
  number={2},
  pages={129--140},
  year={1996},
  publisher={Wiley Online Library}
}

@article{cronbach1965alpha,
  title={Alpha coefficients for stratified-parallel tests},
  author={Cronbach, Lee J and Sch{\"o}nemann, Peter and McKie, Douglas},
  journal={Educational and Psychological Measurement},
  volume={25},
  number={2},
  pages={291--312},
  year={1965},
  publisher={Sage Publications Sage CA: Thousand Oaks, CA}
}

@article{manna2025toefl,
  title={{TOEFL iBT}{\textregistered} technical manual},
  author={Manna, Venessa and Li, Shuhong and Papageorgiou, Spiros and Gu, Lixiong},
  journal={ETS Research Report Series},
  volume={2025},
  number={1},
  year={2025}
}

@article{ikeda2025aligning,
  title={Aligning Scores of Language Proficiency Tests: A Score Concordance Study Between {IELTS} {Academic} and {TOEFL iBT}{\textregistered}},
  author={Ikeda, Naoki and Clark, Tony and Papageorgiou, Spiros and Gu, Lixiong and Ohta, Renka and Blackhurst, Andrew and Bruce, Emma},
  journal={ETS Research Report Series},
  volume={2025},
  number={1},
  pages={1--28},
  year={2025},
  doi = {10.64634/tn27a897}
}

@misc{first2025ef,
  title={{EF EPI: EF English} proficiency index: A ranking of 123 countries and regions by {English} skills},
  author={{Education First}},
  year={2025},
  publisher={EF Education First.},
  url = {https://www.ef.com/assetscdn/WIBIwq6RdJvcD9bc8RMd/cefcom-epi-site/reports/2025/ef-epi-2025-english.pdf},
    note        = {Accessed 29 August 2026}

}

@misc{IELTS2026,
  title={Test statistics},
    author = {{IELTS}},
  year={2026},
  url = {https://ielts.org/researchers/our-research/test-statistics},
    note        = {Accessed August 18th, 2026}

}

\appendix

\clearpage

\section*{Appendix}

\section{Eye Movement Representations}
\label{sec:app-Eye-Movement-Representations}
\subsection{Average Fixation Metrics (Avg. Fix.)}
\label{sec:app-fix-sac-measures}
This feature set includes the following \textbf{6 features}.

\begin{enumerate}
    \item \textbf{First Fixation Duration (FF)} the duration of the first fixation on a word. 
    \item \textbf{Gaze Duration (GD)} the time spent from first entering a word to first leaving it. 
    \item \textbf{Total Fixation Duration (TF)} the sum of all the fixation durations on a word. 
    \item \textbf{Go-Past Duration (GP)} the time from first entering a word until proceeding to its right. 
    \item \textbf{Skip Rate (SR)} The percentage of words skipped during first pass reading. 
    \item \textbf{Regression Probability (RP)} 
    the percentage of words with backward saccades during first-pass reading. 

\end{enumerate}

FF, GD, TF and GP are averaged across all words, with skipped words assigned a value of 0. SR and RP are computed over all the textual materials presented to the reader.

\subsection{Syntactic Clusters (S-Clusters)} The S-Clusters feature set consists of \textbf{258 features}. Each feature is one of the 6 reading measures above, averaged for a given part-of-speech tag.  
We use 43 part-of-speech tags, 14 Universal tags \cite{petrov2012universal} and 29 PTB tags \cite{santorini1990part}, extracted with spaCy \citep{honnibal2020spacy}.

\subsection{Word Property Coefficients (WP-Coefs)} 

The WP-Coefs feature set consists of \textbf{24 features}. The features are the coefficients from a linear model $\mathrm{RT} \sim 1 + \mathrm{length} + \mathrm{surprisal} + \mathrm{frequency}$,
fitted for each participant on each of the measures $\mathrm{RT} \in \{FF, GD, TF, GP\}$. We use the same features in logistic regression models to predict word skips and regressions. . 
Surprisal and word frequency values were extracted using \texttt{psycholing-metrics} version 1.1.9 \cite{yoav_meiri_2026_20376826}. 

\subsection{Transitions} 

Each feature represents the number of saccades in each direction between each pair of words in the passage. We include only pairs that had at least one transition for at least one participant.  
As described in \Cref{sec:eval-setting}, in \textbf{MECO L2}, we divide the passages into two parts and train on each part separately. This results in \textbf{39,442 features}, on average, across the two parts. In \textbf{OneStopL2 and OneStop}, there are 6 unique sets of reading materials (one of the three article batches, in one of two text difficulty level sequences), with an average of \textbf{99,372 features}.

As TabPFN-3 has an upper limit of 2,000 features, we keep a fixed number of Transition features from the middle of each paragraph, such that this limit it met. On average, there are 37 transition features per passage for OneStopL2 and OneStop, and 333 features per passage for MECO L2.

\subsection{Word Fixation Metrics (Word Fix.)} 

The Word Fixation Metrics feature set consists of an average of \textbf{1,658 features in MECO L2} across the two subsets of textual materials, and an average of \textbf{11,719 features in OneStopL2 and OneStop} across the 6 textual material subsets. For TabPFN-3, we take into account only the first 2,000 words presented to the reader, which cover all 12 passages in MECO L2 and the first 9--11 passages in OneStop and OneStopL2.

\subsection{Differences from the Features of \texorpdfstring{\citet{berzak-etal-2018-assessing}}{Berzak et al. (2018)}}

The Syntactic Clusters, Word Property Coefficients, Transitions and Word Fixation Metrics feature sets are similar, but not identical to those used in \citet{berzak-etal-2018-assessing}. Differently from \citet{berzak-etal-2018-assessing}, we do not speed-normalize the underlying eye movement measures, as preliminary experiments suggested that raw measures yielded better performance. We add GP, SR and RP to Syntactic Clusters, and regressions to Word Property Coefficients. When computing Word Property Coefficients, Surprisal is obtained using GPT-2 \citep{radford_language_2019} rather than an n-gram model, and frequency counts are from Wordfreq \citep{robyn_speer_2018} rather than from BLLIP-WSJ \cite{bllip2000}, which matched the domain of their textual materials.

\clearpage 

\section{Experimental Setup}

\subsection{Eye Tracking Datasets}
\label{sec:app-data}

We use the following versions of the eyetracking datasets:
\begin{itemize}
    \item MECO L2 - \texttt{v2.1}.
    \item OneStop - \texttt{v1.0.3}.
    \item OneStopL2 - \texttt{v0.1}.
\end{itemize}

\paragraph{MECO L2}
The MECO L2 dataset contains a total of 1,437,848 word tokens for which eye movement data were collected. The dataset comprises 2,628,941 fixations, corresponding to 2,189 fixations per participant.
The L2 readers account for 1,311,713 word tokens and 2,465,736 fixations, corresponding to 2,229 fixations per participant, while the L1 readers account for 126,135 word tokens and 163,205 fixations, or 1,718 fixations per participant. \Cref{tab:meco_language_dist} presents the number of participants from each L1 background.

\begin{table}[ht!]
\centering
\resizebox{0.7\columnwidth}{!}{
\begin{tabular}{lr}
\toprule
L1 & Participants \\
\midrule
Basque & 35 \\
Brazilian Portuguese & 54 \\
Danish & 25 \\
Dutch & 46 \\
Estonian & 58 \\
Finnish & 51 \\
German & 135 \\
Greek & 48 \\
Hebrew & 45 \\
Hindi & 98 \\
Icelandic & 45 \\
Italian & 51 \\
Mandarin & 89 \\
Norwegian & 62 \\
Russian & 96 \\
Serbian & 43 \\
Spanish & 86 \\
Turkish & 39 \\
\hdashline
\addlinespace[4pt]
English (L1) & 95 \\
\bottomrule
\end{tabular}
}
\caption{\textbf{MECO L2 Participants}. Number of participants from each L1 background.}
\label{tab:meco_language_dist}
\end{table}
 
\paragraph{OneStopL2} We exclude the repeated reading portion of OneStopL2 and OneStop. The first reading portion of OneStopL2 contains 1,600,086 word tokens for which eye movement data were collected. The dataset comprises 2,677,486 fixations, corresponding to 9,772 fixations per participant. \Cref{tab:language_dist} presents the number of participants from each L1 background in the two conditions and overall.

\paragraph{OneStop}
The first reading portion of OneStop contains 2,097,963 word tokens for which eye movement data were collected. The dataset comprises 2,027,969 fixations, corresponding to 5,665 fixations per participant.

\begin{table}[ht!]
\centering
\resizebox{0.85\columnwidth}{!}{
\begin{tabular}{lccc}
\toprule
L1 & \makecell{Ordinary\\Reading} & \makecell{Information\\Seeking} & All \\
\midrule
Arabic & 14 & 11 & 25 \\
Chinese & 18 & 15 & 33 \\
French & 16 & 6 & 22 \\
Hebrew & 15 & 17 & 32 \\
Japanese & 13 & 6 & 19 \\
Korean & 18 & 13 & 31 \\
Portuguese & 18 & 17 & 35 \\
Russian & 17 & 17 & 34 \\
Spanish & 17 & 18 & 35 \\
Vietnamese & 5 & 3 & 8 \\
\bottomrule
\end{tabular}}
\caption{\textbf{OneStopL2 Participants}. Number of participants from each L1 background in the reading for comprehension and information seeking regimes.}
\label{tab:language_dist}
\end{table}

\subsection{Data Splits}
\label{sec:app-eval-settings}

As noted in \Cref{sec:proficiency-tests}, external proficiency test scores are not available for all participants. When evaluating EyeScore against these scores, we retain all participants in the training portion of the splits of \Cref{sec:eval-setting}, regardless of score availability, and evaluate on the held-out participants who have a score for the test in question. For predicting scores on external tests, in contrast, we train and evaluate only on participants with a reported score for that test.
Below, we describe dataset-specific evaluation constraints and the corresponding preprocessing and sampling procedures.

\paragraph{OneStop and OneStopL2}

We keep only participants who had no missing passages in their eye movement recordings. This results in the exclusion of one participant from the information seeking portion of OneStopL2, three from the ordinary reading for comprehension portion of OneStopL2, and two from the ordinary reading for comprehension portion of OneStop. In OneStopL2 \texttt{v0.1}, for a given text, there is a difference in the number of participants reading it in each of the two difficulty versions (original vs simplified), unlike in OneStop. Thus, for the prediction of each external proficiency score in OneStopL2, instead of including all L1 participants in the training set of each split, we include only L1 participants matched to L2 participants in that training set. Specifically, each L2 participant is assigned an L1 participant who read the same texts at the same difficulty versions.

\paragraph{OneStopL2 Information Seeking} 
\Cref{sec:eval-setting} reports the training-split statistics for the ordinary reading for comprehension portion of OneStopL2. For the information seeking part, the corresponding averages are 20 L2 and 20 L1 training participants in the Fixed Text regime, and 81 L2 and 81 L1 in the Any Text regime.

\subsection{Hyperparameter Tuning}
\label{sec:app-tuning}

\paragraph{Ridge regression} In the Any Text regime, for MECO L2, we tune the Ridge regression regularization parameter $\alpha$ twice for each participant, once for each half of their texts, using all other participants' other texts via leave-one-participant-out cross validation. For OneStop, we tune $\alpha$ via leave-one-participant-out cross validation.

In the Fixed Text regime, for MECO L2, we again tune $\alpha$ twice for each participant - once for each half of their texts, this time using all other participants' same texts via leave-one-participant-out cross validation. For OneStop, we tune $\alpha$ for each participant using all other participants in their textual material group via leave-one-participant-out cross validation.

\paragraph{LightGBM} In both regimes, we use 5-fold cross validation. The hyperparameter grid search comprised maximum leaf counts of 15, 31, and 63 per tree, learning rates of 0.03 and 0.1, and feature-sampling fractions of 0.3 and 1.0, where 1.0 indicates that all features were considered. 

\paragraph{TabPFN-3} we use the pretrained model, without any hyperparameter tuning.

\clearpage

\section{The Information Seeking Reading Regime}
\label{sec:app-info-seeking}

\begin{table}[ht!]
\centering
 \small
\begin{tabular}{lcc|cc}
\toprule
\multicolumn{1}{c}{} & \multicolumn{2}{c}{LexTALE} & \multicolumn{2}{c}{Michigan} \\
\textbf{Features} & Fixed & Any & Fixed & Any \\
\midrule
WPM & $0.56$ & $\mathbf{0.59}\phantom{^{*}}$ & $0.45\phantom{^{*}}$ & $0.44$ \\
\midrule
Avg. Fix. & $0.49$ & $0.49^{*}$ & $0.47\phantom{^{*}}$ & $0.44$ \\
S-Clusters & $0.50$ & $0.50^{*}$ & $0.47\phantom{^{*}}$ & $\mathbf{0.45}$ \\
WP-Coefs & $0.48$ & $0.51\phantom{^{*}}$ & $0.43\phantom{^{*}}$ & $0.43$ \\
\midrule
Transitions & $0.58$ & - & $0.49\phantom{^{*}}$ & - \\
Word Fix. & $\mathbf{0.60}$ & - & $\mathbf{0.52}^{*}$ & - \\
\bottomrule
\end{tabular}
\caption{\textbf{EyeScore}. Pearson $r$ correlations of EyeScore with standard English proficiency tests in the \textbf{OneStopL2 information seeking} regime.}
\label{tab:eyescore_seen_unseen_hunting_bootstrap}
\end{table}

\begin{table}[ht!]
\makeatletter
\@ifundefined{hdashline}
  {\providecommand{\meanbaselinerule}{\addlinespace[2pt]\midrule\addlinespace[2pt]}}
  {\@ifundefined{dashlinedash}{}{\setlength\dashlinedash{2pt}\setlength\dashlinegap{2pt}}%
   \providecommand{\meanbaselinerule}{\addlinespace[2pt]\hdashline\addlinespace[2pt]}}
\makeatother
\centering
\resizebox{\columnwidth}{!}{
\begin{tabular}{lcc|cc|cc|cc}
\toprule
\multicolumn{1}{l}{} & \multicolumn{4}{c}{LexTALE} & \multicolumn{4}{c}{Michigan} \\ \cmidrule{2-9}
\multicolumn{1}{l}{} & \multicolumn{2}{c}{Fixed} & \multicolumn{2}{c}{Any} & \multicolumn{2}{c}{Fixed} & \multicolumn{2}{c}{Any} \\ \cmidrule{2-9}
\textbf{Features} & $r$ & MAE & $r$ & MAE & $r$ & MAE & $r$ & MAE \\
\midrule
Avg. Train & $-0.09\phantom{^{*}}$ & $11.27$ & $0.00$ & $10.93\phantom{^{*}}$ & $-0.09\phantom{^{***}}$ & $10.81\phantom{^{**}}$ & $0.00\phantom{^{**}}$ & $10.40$ \\
\meanbaselinerule
WPM & $0.53\phantom{^{*}}$ & $9.95$ & $0.52$ & $10.21\phantom{^{*}}$ & $0.29\phantom{^{***}}$ & $8.90\phantom{^{**}}$ & $0.38\phantom{^{**}}$ & $8.46$ \\
\midrule
Avg. Fix. & $0.53\phantom{^{*}}$ & $10.04$ & $\mathbf{0.60}$ & $9.05^{*}$ & $0.36\phantom{^{***}}$ & $8.81\phantom{^{**}}$ & $\mathbf{0.55}^{**}$ & $\mathbf{7.95}$ \\
S-Clusters & $0.50\phantom{^{*}}$ & $9.94$ & $0.59$ & $\mathbf{8.96}^{*}$ & $0.40\phantom{^{***}}$ & $8.39\phantom{^{**}}$ & $0.53^{**}$ & $8.05$ \\
WP-Coefs & $0.43^{*}$ & $11.01$ & $0.57$ & $9.58\phantom{^{*}}$ & $0.33\phantom{^{***}}$ & $8.90\phantom{^{**}}$ & $0.48\phantom{^{**}}$ & $8.70$ \\
\midrule
Transitions & $\mathbf{0.53}\phantom{^{*}}$ & $\mathbf{9.72}$ & \phantom{0.}-\phantom{00} & \phantom{0.}-\phantom{00} & $0.42^{*}\phantom{^{**}}$ & $7.98^{**}$ & \phantom{0.}-\phantom{00} & \phantom{0.}-\phantom{00} \\
Word Fix. & $0.51\phantom{^{*}}$ & $9.77$ & \phantom{0.}-\phantom{00} & \phantom{0.}-\phantom{00} & $\mathbf{0.53}^{***}$ & $\mathbf{7.76}^{**}$ & \phantom{0.}-\phantom{00} & \phantom{0.}-\phantom{00} \\
\bottomrule
\end{tabular}
}
\caption{\textbf{Prediction of standard English proficiency test scores} with Ridge regression in the \textbf{OneStopL2 information seeking} regime.}
\label{tab:predictions-onestop-hunting}
\end{table}

\twocolumn[{%
\section{Prediction of External Proficiency Tests with LightGBM and TabPFN-3}
\label{sec:app-prediction-more-models}
\vspace{1em}

\definecolor{diffbetter}{rgb}{0.00,0.45,0.70}
\definecolor{diffworse}{rgb}{0.70,0.00,0.00}
\makeatletter
\@ifundefined{hdashline}
  {\providecommand{\meanbaselinerule}{\addlinespace[2pt]\midrule\addlinespace[2pt]}}
  {\@ifundefined{dashlinedash}{}{\setlength\dashlinedash{2pt}\setlength\dashlinegap{2pt}}%
   \providecommand{\meanbaselinerule}{\addlinespace[2pt]\hdashline\addlinespace[2pt]}}
\makeatother
\centering
\resizebox{\ifdim\width>\textwidth\textwidth\else\width\fi}{!}{
\begin{tabular}{ll@{\hspace{4pt}}l|l@{\hspace{4pt}}l|l@{\hspace{4pt}}l|l@{\hspace{4pt}}l}
\toprule
\multicolumn{1}{l}{} & \multicolumn{4}{c}{LexTALE} & \multicolumn{4}{c}{Composite} \\ \cmidrule{2-9}
\multicolumn{1}{l}{} & \multicolumn{2}{c}{Fixed} & \multicolumn{2}{c}{Any} & \multicolumn{2}{c}{Fixed} & \multicolumn{2}{c}{Any} \\ \cmidrule{2-9}
\textbf{Features} & $r$ & MAE & $r$ & MAE & $r$ & MAE & $r$ & MAE \\
\midrule
Avg. Train & $0.00$ & $9.92$ & $0.00$ & $9.92$ & $0.00$ & $11.37$ & $0.00$ & $11.37$ \\
\meanbaselinerule
WPM & $0.48_{{\color{diffworse}(-0.00)}}$ & $8.43_{{\color{diffbetter}(-0.10)}}$ & $0.46_{{\color{diffworse}(-0.02)}}$ & $8.62_{{\color{diffworse}(+0.10)}}$ & $0.45_{{\color{diffbetter}(+0.01)}}$ & $9.90_{{\color{diffbetter}(-0.06)}}$ & $0.46_{{\color{diffbetter}(+0.02)}}$ & $9.80_{{\color{diffbetter}(-0.15)}}$ \\
\midrule
Avg. Fix. & $0.54^{***}_{{\color{diffworse}(-0.00)}}$ & $8.14^{*}_{{\color{diffbetter}(-0.02)}}$ & $0.54^{***}_{{\color{diffworse}(-0.00)}}$ & $8.11^{***}_{{\color{diffbetter}(-0.06)}}$ & $0.52^{***}_{{\color{diffworse}(-0.02)}}$ & $9.50^{**}_{{\color{diffworse}(+0.11)}}$ & $\mathbf{0.53}^{***}_{{\color{diffworse}(-0.01)}}$ & $9.51^{*}_{{\color{diffworse}(+0.10)}}$ \\
S-Clusters & $0.58^{***}_{{\color{diffbetter}(+0.03^{*})}}$ & $7.82^{***}_{{\color{diffbetter}(-0.28^{***})}}$ & $\mathbf{0.55}^{***}_{{\color{diffbetter}(+0.00)}}$ & $8.06^{***}_{{\color{diffbetter}(-0.09)}}$ & $0.57^{***}_{{\color{diffbetter}(+0.01)}}$ & $9.21^{***}_{{\color{diffbetter}(-0.05)}}$ & $\mathbf{0.53}^{***}_{{\color{diffworse}(-0.01)}}$ & $\mathbf{9.46}^{**}_{{\color{diffworse}(+0.07)}}$ \\
WP-Coefs & $0.54^{***}_{{\color{diffworse}(-0.02)}}$ & $8.11^{**}_{{\color{diffbetter}(-0.01)}}$ & $\mathbf{0.55}^{***}_{{\color{diffworse}(-0.00)}}$ & $\mathbf{8.05}^{***}_{{\color{diffbetter}(-0.09)}}$ & $0.53^{***}_{{\color{diffworse}(-0.02)}}$ & $9.44^{***}_{{\color{diffworse}(+0.14)}}$ & $\mathbf{0.53}^{***}_{{\color{diffworse}(-0.01)}}$ & $9.57_{{\color{diffworse}(+0.17)}}$ \\
\midrule
Transitions & $0.59^{***}_{{\color{diffbetter}(+0.04^{***})}}$ & $7.74^{***}_{{\color{diffbetter}(-0.34^{***})}}$ & \phantom{0.}-\phantom{00} & \phantom{0.}-\phantom{00} & $0.53^{***}_{{\color{diffworse}(-0.02^{*})}}$ & $9.34^{***}_{{\color{diffbetter}(-0.03)}}$ & \phantom{0.}-\phantom{00} & \phantom{0.}-\phantom{00} \\
Word Fix. & $\mathbf{0.63}^{***}_{{\color{diffbetter}(+0.01)}}$ & $\mathbf{7.45}^{***}_{{\color{diffbetter}(-0.14)}}$ & \phantom{0.}-\phantom{00} & \phantom{0.}-\phantom{00} & $\mathbf{0.61}^{***}_{{\color{diffbetter}(+0.02)}}$ & $\mathbf{8.76}^{***}_{{\color{diffbetter}(-0.10)}}$ & \phantom{0.}-\phantom{00} & \phantom{0.}-\phantom{00} \\
\bottomrule
\end{tabular}
}
\captionof{table}{\textbf{Prediction of standard English proficiency test scores}  with \textbf{LightGBM} on \textbf{MECO L2}.
Statistically significant differences from the WPM baseline are marked in the superscript. Differences from the Ridge regression results (\Cref{tab:predictions_bootstrap}) are marked in the subscript in \textcolor{diffbetter}{blue} when better than Ridge regression and \textcolor{diffworse}{red} when worse. Statistically significant differences (bootstrap test) are marked with `$^{*}$' $p<0.05$, `$^{**}$' $p<0.01$, `$^{***}$' $p<0.001$.}
\label{tab:predictions_bootstrap_lgbm_holdout_meco}

\vspace{2em}

\definecolor{diffbetter}{rgb}{0.00,0.45,0.70}
\definecolor{diffworse}{rgb}{0.70,0.00,0.00}
\makeatletter
\@ifundefined{hdashline}
  {\providecommand{\meanbaselinerule}{\addlinespace[2pt]\midrule\addlinespace[2pt]}}
  {\@ifundefined{dashlinedash}{}{\setlength\dashlinedash{2pt}\setlength\dashlinegap{2pt}}%
   \providecommand{\meanbaselinerule}{\addlinespace[2pt]\hdashline\addlinespace[2pt]}}
\makeatother
\centering
\resizebox{\ifdim\width>\textwidth\textwidth\else\width\fi}{!}{
\begin{tabular}{ll@{\hspace{4pt}}l|l@{\hspace{4pt}}l|l@{\hspace{4pt}}l|l@{\hspace{4pt}}l}
\toprule
\multicolumn{1}{l}{} & \multicolumn{4}{c}{LexTALE} & \multicolumn{4}{c}{Michigan} \\ \cmidrule{2-9}
\multicolumn{1}{l}{} & \multicolumn{2}{c}{Fixed} & \multicolumn{2}{c}{Any} & \multicolumn{2}{c}{Fixed} & \multicolumn{2}{c}{Any} \\ \cmidrule{2-9}
\textbf{Features} & $r$ & MAE & $r$ & MAE & $r$ & MAE & $r$ & MAE \\
\midrule
Avg. Train & $-0.06$ & $10.32$ & $-0.02$ & $10.25$ & $-0.07$ & $7.52$ & $-0.01$ & $7.40$ \\
\meanbaselinerule
WPM & $0.19_{{\color{diffworse}(-0.23^{**})}}$ & $11.26_{{\color{diffworse}(+1.06^{*})}}$ & $0.44_{{\color{diffworse}(-0.04)}}$ & $9.56_{{\color{diffbetter}(-0.06)}}$ & $0.13_{{\color{diffworse}(-0.24^{***})}}$ & $7.54_{{\color{diffworse}(+1.12^{***})}}$ & $0.45_{{\color{diffworse}(-0.03)}}$ & $6.37_{{\color{diffworse}(+0.23)}}$ \\
\midrule
Avg. Fix. & $0.26_{{\color{diffworse}(-0.21^{**})}}$ & $10.94_{{\color{diffworse}(+0.94)}}$ & $0.45_{{\color{diffworse}(-0.09^{*})}}$ & $9.85_{{\color{diffworse}(+0.81^{*})}}$ & $0.16_{{\color{diffworse}(-0.30^{***})}}$ & $7.44_{{\color{diffworse}(+1.10^{**})}}$ & $0.49_{{\color{diffworse}(-0.07^{**})}}$ & $6.40_{{\color{diffworse}(+0.53^{*})}}$ \\
S-Clusters & $0.36^{**}_{{\color{diffworse}(-0.13^{**})}}$ & $10.19^{*}_{{\color{diffworse}(+0.44)}}$ & $\mathbf{0.53}_{{\color{diffworse}(-0.04)}}$ & $9.13_{{\color{diffworse}(+0.46)}}$ & $0.24_{{\color{diffworse}(-0.23^{***})}}$ & $7.28_{{\color{diffworse}(+1.03^{***})}}$ & $\mathbf{0.50}_{{\color{diffworse}(-0.07^{*})}}$ & $\mathbf{6.13}_{{\color{diffworse}(+0.31)}}$ \\
WP-Coefs & $0.34^{*}_{{\color{diffworse}(-0.14^{*})}}$ & $10.30_{{\color{diffworse}(+0.66)}}$ & $0.51_{{\color{diffworse}(-0.07^{*})}}$ & $\mathbf{8.91}_{{\color{diffworse}(+0.33)}}$ & $0.28^{*}_{{\color{diffworse}(-0.18^{*})}}$ & $7.14_{{\color{diffworse}(+0.77^{*})}}$ & $0.48_{{\color{diffworse}(-0.08)}}$ & $6.13_{{\color{diffworse}(+0.22)}}$ \\
\midrule
Transitions & $0.45^{***}_{{\color{diffworse}(-0.06)}}$ & $10.19_{{\color{diffworse}(+0.41)}}$ & \phantom{0.}-\phantom{00} & \phantom{0.}-\phantom{00} & $0.29_{{\color{diffworse}(-0.17^{**})}}$ & $6.96_{{\color{diffworse}(+0.89^{***})}}$ & \phantom{0.}-\phantom{00} & \phantom{0.}-\phantom{00} \\
Word Fix. & $\mathbf{0.46}^{***}_{{\color{diffworse}(-0.09^{*})}}$ & $\mathbf{9.68}^{**}_{{\color{diffworse}(+0.82)}}$ & \phantom{0.}-\phantom{00} & \phantom{0.}-\phantom{00} & $\mathbf{0.49}^{***}_{{\color{diffworse}(-0.03)}}$ & $\mathbf{5.95}^{***}_{{\color{diffworse}(+0.10)}}$ & \phantom{0.}-\phantom{00} & \phantom{0.}-\phantom{00} \\
\bottomrule
\end{tabular}
}
\captionof{table}{\textbf{Prediction of standard English proficiency test scores} with \textbf{LightGBM} on \textbf{OneStopL2}.}
\label{tab:predictions_bootstrap_lgbm_holdout_onestop}
\vspace{2em}

\definecolor{diffbetter}{rgb}{0.00,0.45,0.70}
\definecolor{diffworse}{rgb}{0.70,0.00,0.00}
\makeatletter
\@ifundefined{hdashline}
  {\providecommand{\meanbaselinerule}{\addlinespace[2pt]\midrule\addlinespace[2pt]}}
  {\@ifundefined{dashlinedash}{}{\setlength\dashlinedash{2pt}\setlength\dashlinegap{2pt}}%
   \providecommand{\meanbaselinerule}{\addlinespace[2pt]\hdashline\addlinespace[2pt]}}
\makeatother
\centering
\resizebox{\ifdim\width>\textwidth\textwidth\else\width\fi}{!}{
\begin{tabular}{ll@{\hspace{4pt}}l|l@{\hspace{4pt}}l|l@{\hspace{4pt}}l|l@{\hspace{4pt}}l}
\toprule
\multicolumn{1}{l}{} & \multicolumn{4}{c}{LexTALE} & \multicolumn{4}{c}{Composite} \\ \cmidrule{2-9}
\multicolumn{1}{l}{} & \multicolumn{2}{c}{Fixed} & \multicolumn{2}{c}{Any} & \multicolumn{2}{c}{Fixed} & \multicolumn{2}{c}{Any} \\ \cmidrule{2-9}
\textbf{Features} & $r$ & MAE & $r$ & MAE & $r$ & MAE & $r$ & MAE \\
\midrule
Avg. Train & $0.00$ & $9.92$ & $0.00$ & $9.92$ & $0.00$ & $11.37$ & $0.00$ & $11.37$ \\
\meanbaselinerule
WPM & $0.49_{{\color{diffbetter}(+0.01^{*})}}$ & $8.41_{{\color{diffbetter}(-0.12^{**})}}$ & $0.49_{{\color{diffbetter}(+0.01^{*})}}$ & $8.42_{{\color{diffbetter}(-0.10^{*})}}$ & $0.48_{{\color{diffbetter}(+0.03^{***})}}$ & $9.71_{{\color{diffbetter}(-0.25^{***})}}$ & $0.48_{{\color{diffbetter}(+0.03^{***})}}$ & $9.72_{{\color{diffbetter}(-0.24^{***})}}$ \\
\midrule
Avg. Fix. & $0.57^{***}_{{\color{diffbetter}(+0.03^{***})}}$ & $7.95^{***}_{{\color{diffbetter}(-0.22^{**})}}$ & $0.57^{***}_{{\color{diffbetter}(+0.03^{***})}}$ & $7.93^{***}_{{\color{diffbetter}(-0.24^{***})}}$ & $0.56^{***}_{{\color{diffbetter}(+0.02^{*})}}$ & $9.16^{***}_{{\color{diffbetter}(-0.23^{***})}}$ & $\mathbf{0.56}^{***}_{{\color{diffbetter}(+0.02^{*})}}$ & $\mathbf{9.17}^{***}_{{\color{diffbetter}(-0.24^{***})}}$ \\
S-Clusters & $0.59^{***}_{{\color{diffbetter}(+0.04^{***})}}$ & $7.77^{***}_{{\color{diffbetter}(-0.33^{***})}}$ & $\mathbf{0.58}^{***}_{{\color{diffbetter}(+0.03^{***})}}$ & $\mathbf{7.87}^{***}_{{\color{diffbetter}(-0.27^{***})}}$ & $0.57^{***}_{{\color{diffbetter}(+0.02^{*})}}$ & $9.11^{***}_{{\color{diffbetter}(-0.15^{*})}}$ & $0.55^{***}_{{\color{diffbetter}(+0.01^{*})}}$ & $9.26^{***}_{{\color{diffbetter}(-0.12^{*})}}$ \\
WP-Coefs & $0.58^{***}_{{\color{diffbetter}(+0.03^{***})}}$ & $7.87^{***}_{{\color{diffbetter}(-0.25^{***})}}$ & $\mathbf{0.58}^{***}_{{\color{diffbetter}(+0.02^{**})}}$ & $7.91^{***}_{{\color{diffbetter}(-0.23^{***})}}$ & $0.55^{***}_{{\color{diffbetter}(+0.00)}}$ & $9.27^{***}_{{\color{diffbetter}(-0.03)}}$ & $0.55^{***}_{{\color{diffbetter}(+0.01)}}$ & $9.35^{***}_{{\color{diffbetter}(-0.06)}}$ \\
\midrule
Transitions & $0.54^{*}_{{\color{diffworse}(-0.01)}}$ & $8.23_{{\color{diffworse}(+0.15^{*})}}$ & \phantom{0.}-\phantom{00} & \phantom{0.}-\phantom{00} & $0.51_{{\color{diffworse}(-0.05^{***})}}$ & $9.63_{{\color{diffworse}(+0.26^{***})}}$ & \phantom{0.}-\phantom{00} & \phantom{0.}-\phantom{00} \\
Word Fix. & $\mathbf{0.63}^{***}_{{\color{diffbetter}(+0.02)}}$ & $\mathbf{7.40}^{***}_{{\color{diffbetter}(-0.19^{*})}}$ & \phantom{0.}-\phantom{00} & \phantom{0.}-\phantom{00} & $\mathbf{0.60}^{***}_{{\color{diffbetter}(+0.01)}}$ & $\mathbf{8.77}^{***}_{{\color{diffbetter}(-0.09)}}$ & \phantom{0.}-\phantom{00} & \phantom{0.}-\phantom{00} \\
\bottomrule
\end{tabular}
}
\captionof{table}{\textbf{Prediction of standard English proficiency test scores} with \textbf{TabPFN-3} on \textbf{MECO L2}.}
\label{tab:predictions_bootstrap_tabpfn_meco}
}]

\twocolumn[{%
\definecolor{diffbetter}{rgb}{0.00,0.45,0.70}
\definecolor{diffworse}{rgb}{0.70,0.00,0.00}
\makeatletter
\@ifundefined{hdashline}
  {\providecommand{\meanbaselinerule}{\addlinespace[2pt]\midrule\addlinespace[2pt]}}
  {\@ifundefined{dashlinedash}{}{\setlength\dashlinedash{2pt}\setlength\dashlinegap{2pt}}%
   \providecommand{\meanbaselinerule}{\addlinespace[2pt]\hdashline\addlinespace[2pt]}}
\makeatother
\centering
\resizebox{\ifdim\width>\textwidth\textwidth\else\width\fi}{!}{
\begin{tabular}{ll@{\hspace{4pt}}l|l@{\hspace{4pt}}l|l@{\hspace{4pt}}l|l@{\hspace{4pt}}l}
\toprule
\multicolumn{1}{l}{} & \multicolumn{4}{c}{LexTALE} & \multicolumn{4}{c}{Michigan} \\ \cmidrule{2-9}
\multicolumn{1}{l}{} & \multicolumn{2}{c}{Fixed} & \multicolumn{2}{c}{Any} & \multicolumn{2}{c}{Fixed} & \multicolumn{2}{c}{Any} \\ \cmidrule{2-9}
\textbf{Features} & $r$ & MAE & $r$ & MAE & $r$ & MAE & $r$ & MAE \\
\midrule
Avg. Train & $-0.06$ & $10.32$ & $-0.02$ & $10.25$ & $-0.07$ & $7.52$ & $-0.01$ & $7.40$ \\
\meanbaselinerule
WPM & $0.42_{{\color{diffbetter}(+0.00)}}$ & $9.80_{{\color{diffbetter}(-0.40)}}$ & $0.49_{{\color{diffbetter}(+0.01)}}$ & $9.43_{{\color{diffbetter}(-0.19)}}$ & $0.41_{{\color{diffbetter}(+0.04)}}$ & $6.66_{{\color{diffworse}(+0.24)}}$ & $0.49_{{\color{diffbetter}(+0.02)}}$ & $6.22_{{\color{diffworse}(+0.08)}}$ \\
\midrule
Avg. Fix. & $0.46_{{\color{diffworse}(-0.01)}}$ & $9.79_{{\color{diffbetter}(-0.22)}}$ & $0.53_{{\color{diffworse}(-0.01)}}$ & $9.10_{{\color{diffworse}(+0.06)}}$ & $0.47_{{\color{diffbetter}(+0.02)}}$ & $6.79_{{\color{diffworse}(+0.45^{*})}}$ & $0.56^{*}_{{\color{diffworse}(-0.00)}}$ & $6.04_{{\color{diffworse}(+0.18)}}$ \\
S-Clusters & $0.50_{{\color{diffbetter}(+0.01)}}$ & $8.99_{{\color{diffbetter}(-0.75^{*})}}$ & $0.54_{{\color{diffworse}(-0.03)}}$ & $9.05_{{\color{diffworse}(+0.38)}}$ & $0.37_{{\color{diffworse}(-0.10^{**})}}$ & $7.14_{{\color{diffworse}(+0.89^{***})}}$ & $\mathbf{0.57}_{{\color{diffbetter}(+0.00)}}$ & $\mathbf{5.91}_{{\color{diffworse}(+0.10)}}$ \\
WP-Coefs & $0.50_{{\color{diffbetter}(+0.02)}}$ & $9.06_{{\color{diffbetter}(-0.59)}}$ & $\mathbf{0.58}^{*}_{{\color{diffworse}(-0.00)}}$ & $\mathbf{8.36}^{**}_{{\color{diffbetter}(-0.22)}}$ & $0.42_{{\color{diffworse}(-0.04)}}$ & $7.00_{{\color{diffworse}(+0.64^{*})}}$ & $0.55_{{\color{diffworse}(-0.01)}}$ & $6.01_{{\color{diffworse}(+0.11)}}$ \\
\midrule
Transitions & $0.51_{{\color{diffbetter}(+0.01)}}$ & $9.49_{{\color{diffbetter}(-0.29)}}$ & \phantom{0.}-\phantom{00} & \phantom{0.}-\phantom{00} & $0.49^{*}_{{\color{diffbetter}(+0.02)}}$ & $6.31_{{\color{diffworse}(+0.23)}}$ & \phantom{0.}-\phantom{00} & \phantom{0.}-\phantom{00} \\
Word Fix. & $\mathbf{0.56}^{***}_{{\color{diffworse}(-0.00)}}$ & $\mathbf{8.96}^{*}_{{\color{diffworse}(+0.10)}}$ & \phantom{0.}-\phantom{00} & \phantom{0.}-\phantom{00} & $\mathbf{0.55}^{***}_{{\color{diffbetter}(+0.03)}}$ & $\mathbf{5.87}^{**}_{{\color{diffworse}(+0.01)}}$ & \phantom{0.}-\phantom{00} & \phantom{0.}-\phantom{00} \\
\bottomrule
\end{tabular}
}
\captionof{table}{\textbf{Prediction of standard English proficiency test scores} with \textbf{TabPFN-3} on \textbf{OneStopL2}.}
\label{tab:predictions_bootstrap_tabpfn_onestop}
\vspace{2em}

\section{Measuring and Mitigating L1 Bias}
\label{sec:app-debias}

\subsection{L1 Bias Across Feature Sets and Datasets}\label{sec:app-bias-obtain-methods-data}

\vspace{1em}
\begin{center}
    \includegraphics[width=1\textwidth]{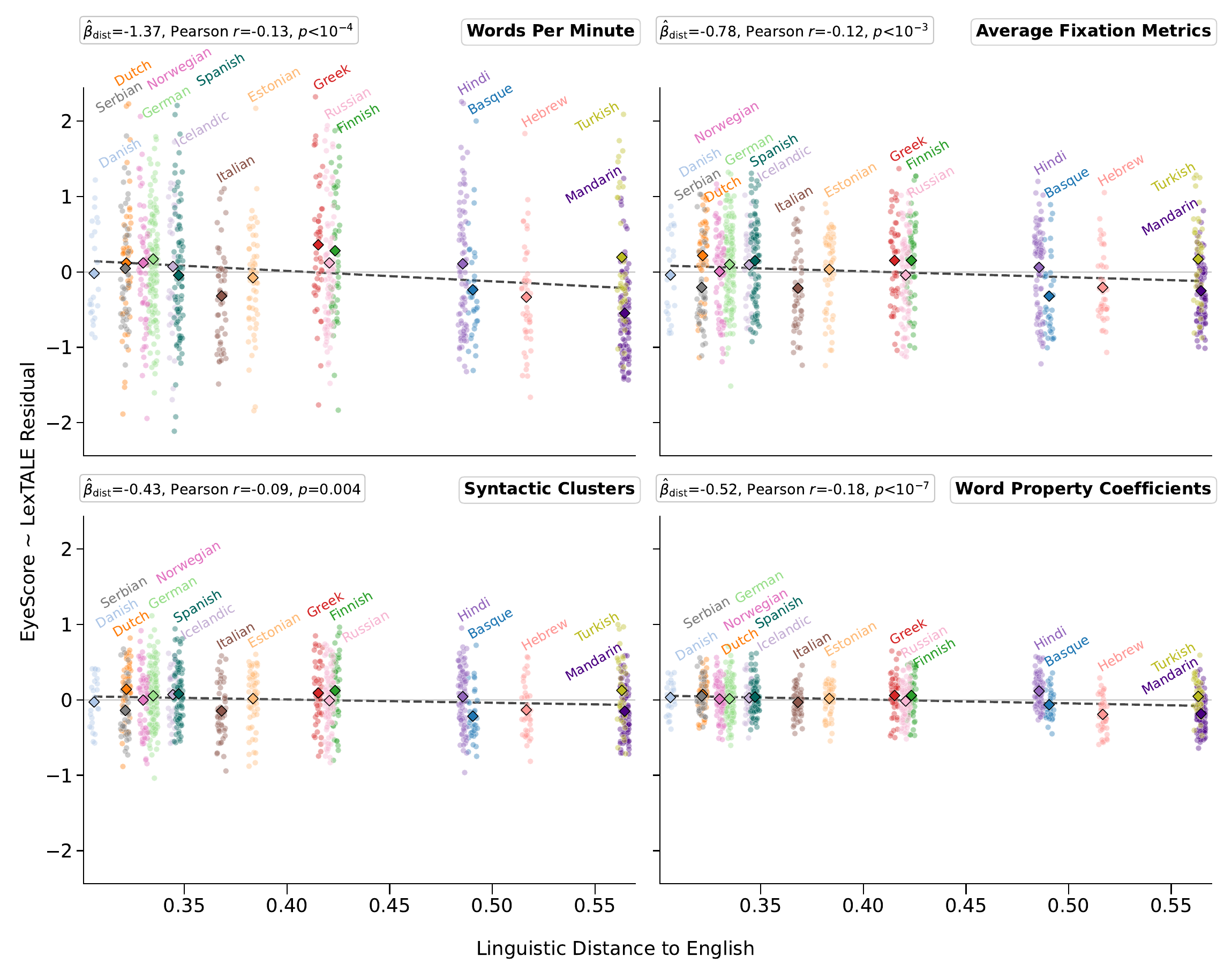}
    \captionof{figure}{\textbf{L1-English Proximity Bias in EyeScore}, \textbf{MECO L2}. Residuals of an \textbf{EyeScore} $\sim$ \textbf{LexTALE} regression as a function of the linguistic distance of the participant's L1 to English, in the Any Text regime.}
    \label{fig:residual_regression_meco}
\end{center}
\vspace{1em}
}]

\begin{figure*}[ht!]
    \centering
    \includegraphics[width=1\textwidth]{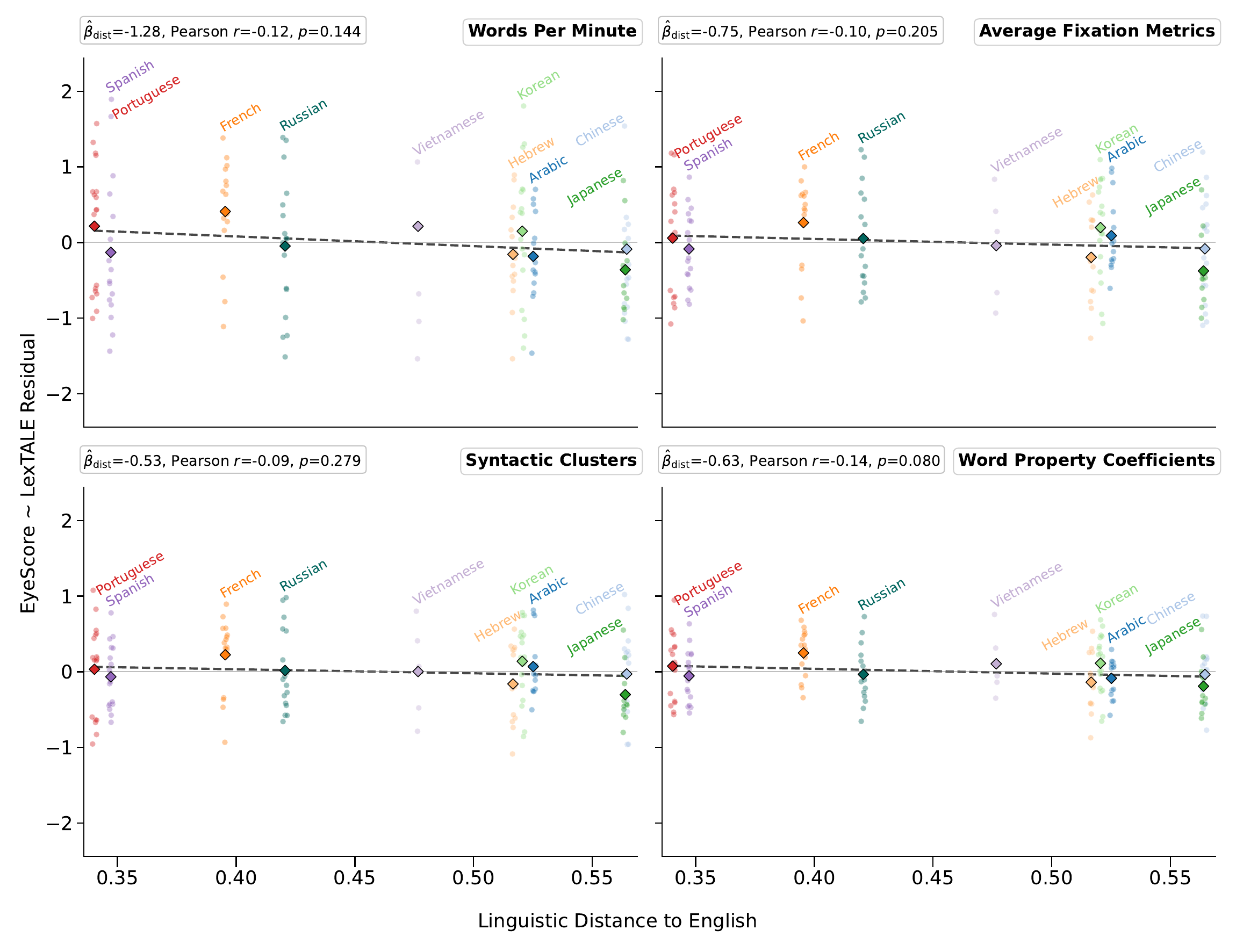}
    \caption{\textbf{L1-English Proximity Bias in EyeScore},  \textbf{OneStopL2}. Residuals of an \textbf{EyeScore} $\sim$ \textbf{LexTALE} regression as a function of the linguistic distance of the participant's L1 to English, in the Any Text regime.}
    \label{fig:residual_regression_onestop}
\end{figure*}

\begin{figure*}[ht!]
    \centering
    \includegraphics[width=1\textwidth]{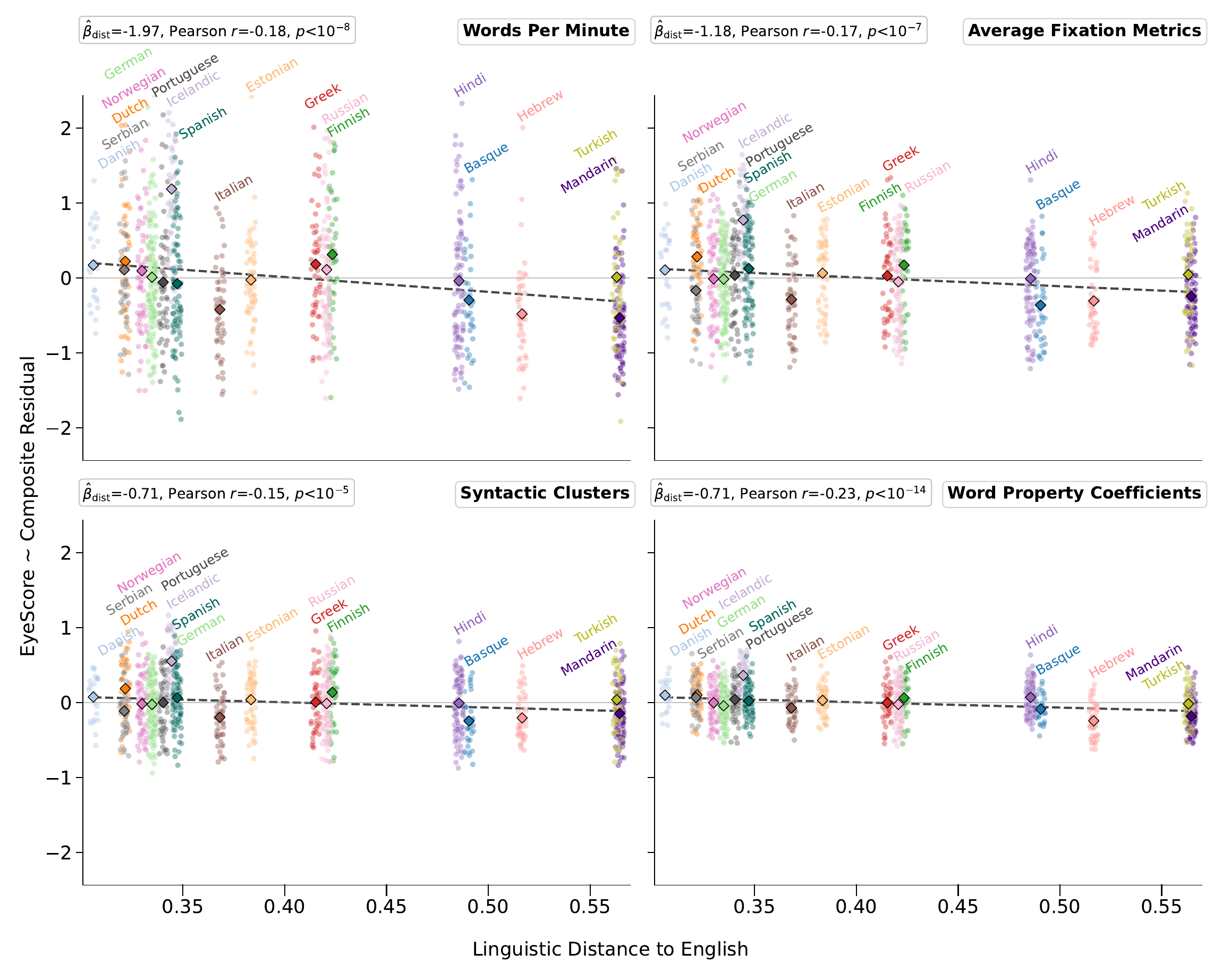}
    \caption{\textbf{L1-English Proximity Bias in EyeScore},  \textbf{MECO L2}. Residuals of an \textbf{EyeScore} $\sim$ \textbf{Composite} regression as a function of the linguistic distance of the participant's L1 to English, in the Any Text regime.}
    \label{fig:residual_regression_meco_composite}
\end{figure*}

\begin{figure*}[ht!]
    \centering
    \includegraphics[width=1\textwidth]{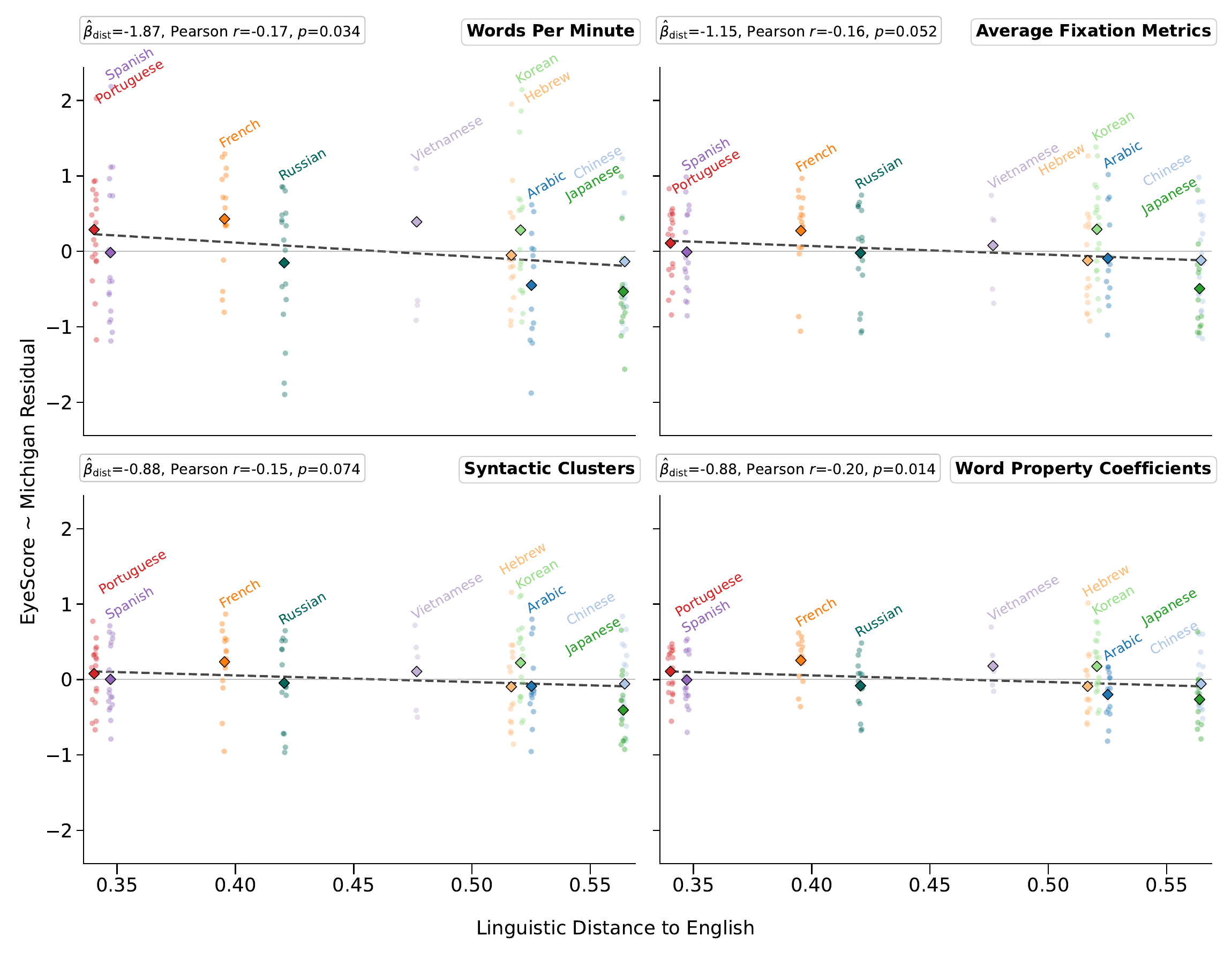}
    \caption{\textbf{L1-English Proximity Bias in EyeScore}, \textbf{OneStopL2}. Residuals of an \textbf{EyeScore} $\sim$ \textbf{Michigan} regression as a function of the linguistic distance of the participant's L1 to English, in the Any Text regime.}
    \label{fig:residual_regression_onestop_michigan}
\end{figure*}

\twocolumn[{%
\subsection{L1 Bias: Pearson $\mathbf{r}$}\label{sec:app-bias-bias}
\vspace{0.4em}

\centering
\setlength{\aboverulesep}{0pt}
\setlength{\belowrulesep}{0pt}
\renewcommand{\arraystretch}{1.25}
\resizebox{\textwidth}{!}{%
\begin{tabular}{lccc|ccc|ccc|ccc}
\toprule
 & \multicolumn{6}{c}{\textbf{MECO L2}} & \multicolumn{6}{c}{\textbf{OneStopL2}} \\
 & \multicolumn{3}{c}{LexTALE} & \multicolumn{3}{c|}{Composite} & \multicolumn{3}{c}{LexTALE} & \multicolumn{3}{c}{Michigan} \\ \cmidrule{2-13}
 & \multicolumn{1}{c}{EyeScore} & \multicolumn{2}{c|}{$\text{EyeScore}^{\mathrm{db}}$} & \multicolumn{1}{c}{EyeScore} & \multicolumn{2}{c|}{$\text{EyeScore}^{\mathrm{db}}$} & \multicolumn{1}{c}{EyeScore} & \multicolumn{2}{c|}{$\text{EyeScore}^{\mathrm{db}}$} & \multicolumn{1}{c}{EyeScore} & \multicolumn{2}{c}{$\text{EyeScore}^{\mathrm{db}}$} \\ \cmidrule{2-13}
\textbf{Features} &  & \textit{Seen L1} & \textit{New L1} &  & \textit{Seen L1} & \textit{New L1} &  & \textit{Seen L1} & \textit{New L1} &  & \textit{Seen L1} & \textit{New L1} \\
\midrule
WPM & $-0.13^{***}$ & $\phantom{-}0.00$ & $-0.04$ & $-0.18^{***}$ & $-0.06\phantom{^{*}}$ & $-0.09^{**}$ & $-0.12$ & $\phantom{-}0.00$ & $\phantom{-}0.00$ & $-0.17^{*}$ & $-0.06$ & $-0.05$ \\
\midrule
Avg. Fix. & $-0.12^{***}$ & $\phantom{-}0.00$ & $-0.02$ & $-0.17^{***}$ & $-0.06\phantom{^{*}}$ & $-0.07^{*}\phantom{^{*}}$ & $-0.10$ & $\phantom{-}0.00$ & $\phantom{-}0.00$ & $-0.16\phantom{^{*}}$ & $-0.06$ & $-0.05$ \\
S-Clusters & $-0.09^{**}\phantom{^{*}}$ & $\phantom{-}0.00$ & $-0.02$ & $-0.15^{***}$ & $-0.06\phantom{^{*}}$ & $-0.08^{*}\phantom{^{*}}$ & $-0.09$ & $\phantom{-}0.00$ & $\phantom{-}0.01$ & $-0.15\phantom{^{*}}$ & $-0.06$ & $-0.05$ \\
WP-Coefs & $-0.18^{***}$ & $-0.01$ & $-0.03$ & $-0.23^{***}$ & $-0.06^{*}$ & $-0.09^{**}$ & $-0.14$ & $\phantom{-}0.00$ & $\phantom{-}0.00$ & $-0.20^{*}$ & $-0.06$ & $-0.05$ \\ \bottomrule
\end{tabular}%
}
\captionof{table}{\textbf{L1-English Proximity Bias in EyeScore and EyeScore\textsuperscript{db}} in the \textbf{Any Text} regime. \textbf{Pearson $\mathbf{r}$} of the model $\mathrm{res}_{p,\mathrm{Test}} \sim d(\text{L1}_p, \text{English})$.}
\label{tab:lang-bias-any-pearson}

\vspace{0.6em}

\subsection{Debiasing Obtained by Composite and Michigan}\label{sec:app-bias-obtain-by}
\vspace{0.4em}

\noindent
\begin{minipage}[t]{0.49\textwidth}
\centering
\resizebox{\linewidth}{!}{%
\begin{tabular}{lccc|ccc}
\toprule
 & \multicolumn{3}{c}{LexTALE} & \multicolumn{3}{c}{Composite} \\ \cmidrule{2-7}
 & \multicolumn{1}{c}{EyeScore} & \multicolumn{2}{c|}{$\text{EyeScore}^{\mathrm{db}}$} & \multicolumn{1}{c}{EyeScore} & \multicolumn{2}{c}{$\text{EyeScore}^{\mathrm{db}}$} \\ \cmidrule{2-7}
\textbf{Features} &  & \textit{Seen L1} & \textit{New L1} &  & \textit{Seen L1} & \textit{New L1} \\
\midrule \cmidrule{1-7}
WPM & $-1.37^{***}$ & $\phantom{-}0.52$ & $\phantom{-}0.30$ & $-1.97^{***}$ & $-0.01$ & $-0.24$ \\
\midrule
Avg. Fix. & $-0.78^{***}$ & $\phantom{-}0.35$ & $\phantom{-}0.35$ & $-1.18^{***}$ & $-0.01$ & $-0.02$ \\
S-Clusters & $-0.43^{**}\phantom{^{*}}$ & $\phantom{-}0.25$ & $\phantom{-}0.23$ & $-0.71^{***}$ & $\phantom{-}0.00$ & $-0.02$ \\
WP-Coefs & $-0.52^{***}$ & $\phantom{-}0.16$ & $\phantom{-}0.13$ & $-0.71^{***}$ & $\phantom{-}0.00$ & $-0.04$ \\ \bottomrule
\end{tabular}%
}
\captionof{table}{\textbf{L1-English Proximity Bias in EyeScore and EyeScore\textsuperscript{db}} in the Any Text regime, \textbf{MECO L2}. EyeScore\textsuperscript{db} obtained by debiasing EyeScore relative to the \textbf{Composite} test.}
\label{tab:lang-bias-triplet-coef-two_step-two_step-compositedb-full-any-coef}
\end{minipage}
\hfill
\begin{minipage}[t]{0.49\textwidth}
\centering
\resizebox{\linewidth}{!}{%
\begin{tabular}{lccc|ccc}
\toprule
 & \multicolumn{3}{c}{LexTALE} & \multicolumn{3}{c}{Michigan} \\ \cmidrule{2-7}
 & \multicolumn{1}{c}{EyeScore} & \multicolumn{2}{c|}{$\text{EyeScore}^{\mathrm{db}}$} & \multicolumn{1}{c}{EyeScore} & \multicolumn{2}{c}{$\text{EyeScore}^{\mathrm{db}}$} \\ \cmidrule{2-7}
\textbf{Features} &  & \textit{Seen L1} & \textit{New L1} &  & \textit{Seen L1} & \textit{New L1} \\
\midrule \cmidrule{1-7}
WPM & $-1.28$ & $\phantom{-}0.64$ & $\phantom{-}0.61$ & $-1.87^{*}$ & $\phantom{-}0.03$ & $\phantom{-}0.06$ \\
\midrule
Avg. Fix. & $-0.75$ & $\phantom{-}0.41$ & $\phantom{-}0.47$ & $-1.15\phantom{^{*}}$ & $\phantom{-}0.02$ & $\phantom{-}0.09$ \\
S-Clusters & $-0.53$ & $\phantom{-}0.34$ & $\phantom{-}0.41$ & $-0.88\phantom{^{*}}$ & $\phantom{-}0.00$ & $\phantom{-}0.08$ \\
WP-Coefs & $-0.63$ & $\phantom{-}0.23$ & $\phantom{-}0.26$ & $-0.88^{*}$ & $\phantom{-}0.00$ & $\phantom{-}0.03$ \\ \bottomrule
\end{tabular}%
}
\captionof{table}{\textbf{L1-English Proximity Bias in EyeScore and EyeScore\textsuperscript{db}} in the Any Text regime, \textbf{OneStopL2}. EyeScore\textsuperscript{db} obtained by debiasing EyeScore relative to the \textbf{Michigan} test.}
\label{tab:lang-bias-triplet-coef-two_step-two_step-michigandb-full-any-coef}
\end{minipage}
\vspace{0.6em}

\subsection{L1 Bias in the Fixed Text and Information Seeking Regimes}
\label{sec:app-bias-additional-results}
\vspace{0.4em}

\begingroup
\centering
\setlength{\aboverulesep}{0pt}
\setlength{\belowrulesep}{0pt}
\renewcommand{\arraystretch}{1.25}
\resizebox{\textwidth}{!}{%
\begin{tabular}{lccc|ccc|ccc|ccc}
\toprule
 & \multicolumn{6}{c}{\textbf{MECO L2}} & \multicolumn{6}{c}{\textbf{OneStopL2}} \\
 & \multicolumn{3}{c}{LexTALE} & \multicolumn{3}{c|}{Composite} & \multicolumn{3}{c}{LexTALE} & \multicolumn{3}{c}{Michigan} \\ \cmidrule{2-13}
 & \multicolumn{1}{c}{EyeScore} & \multicolumn{2}{c|}{$\text{EyeScore}^{\mathrm{db}}$} & \multicolumn{1}{c}{EyeScore} & \multicolumn{2}{c|}{$\text{EyeScore}^{\mathrm{db}}$} & \multicolumn{1}{c}{EyeScore} & \multicolumn{2}{c|}{$\text{EyeScore}^{\mathrm{db}}$} & \multicolumn{1}{c}{EyeScore} & \multicolumn{2}{c}{$\text{EyeScore}^{\mathrm{db}}$} \\ \cmidrule{2-13}
\textbf{Features} &  & \textit{Seen L1} & \textit{New L1} &  & \textit{Seen L1} & \textit{New L1} &  & \textit{Seen L1} & \textit{New L1} &  & \textit{Seen L1} & \textit{New L1} \\
\midrule
WPM & $-1.37^{***}$ & $-0.05\phantom{^{***}}$ & $-0.43\phantom{^{***}}$ & $-1.96^{***}$ & $-0.60\phantom{^{***}}$ & $-0.97^{**}\phantom{^{*}}$ & $-1.57\phantom{^{*}}$ & $-0.11$ & $-0.01$ & $-2.08^{*}\phantom{^{*}}$ & $-0.70$ & $-0.52$ \\
\midrule
Avg. Fix. & $-0.78^{***}$ & $-0.03\phantom{^{***}}$ & $-0.15\phantom{^{***}}$ & $-1.18^{***}$ & $-0.41\phantom{^{***}}$ & $-0.52^{*}\phantom{^{**}}$ & $-0.69\phantom{^{*}}$ & $\phantom{-}0.07$ & $\phantom{-}0.10$ & $-1.07\phantom{^{**}}$ & $-0.32$ & $-0.26$ \\
S-Clusters & $-0.48^{**}\phantom{^{*}}$ & $-0.02\phantom{^{***}}$ & $-0.11\phantom{^{***}}$ & $-0.77^{***}$ & $-0.29\phantom{^{***}}$ & $-0.39^{**}\phantom{^{*}}$ & $-0.53\phantom{^{*}}$ & $\phantom{-}0.07$ & $\phantom{-}0.08$ & $-0.87\phantom{^{**}}$ & $-0.27$ & $-0.23$ \\
WP-Coefs & $-0.49^{***}$ & $-0.02\phantom{^{***}}$ & $-0.11\phantom{^{***}}$ & $-0.70^{***}$ & $-0.21^{*}\phantom{^{**}}$ & $-0.29^{**}\phantom{^{*}}$ & $-0.73^{*}$ & $\phantom{-}0.02$ & $\phantom{-}0.05$ & $-0.97^{**}$ & $-0.20$ & $-0.17$ \\
\midrule
Transitions & $-0.05^{***}$ & $-0.07^{***}$ & $-0.09^{***}$ & $-0.09^{***}$ & $-0.11^{***}$ & $-0.12^{***}$ & $-0.14\phantom{^{*}}$ & $-0.13$ & $-0.12$ & $-0.22^{*}\phantom{^{*}}$ & $-0.23$ & $-0.22$ \\
Word Fix. & $-0.50^{***}$ & $-0.02\phantom{^{***}}$ & $-0.11\phantom{^{***}}$ & $-0.70^{***}$ & $-0.20^{*}\phantom{^{**}}$ & $-0.29^{**}\phantom{^{*}}$ & $-0.44\phantom{^{*}}$ & $-0.05$ & $-0.07$ & $-0.64^{*}\phantom{^{*}}$ & $-0.25$ & $-0.25$ \\ \bottomrule
\end{tabular}%
}
\captionof{table}{\textbf{L1-English Proximity Bias in EyeScore and EyeScore\textsuperscript{db}} in the \textbf{Fixed Text} regime.}
\label{tab:lang-bias-fixed-coef}
\endgroup

\vspace{0.6em}

\begingroup
\centering
\resizebox{0.62\textwidth}{!}{%
\begin{tabular}{lccc|ccc}
\toprule
 & \multicolumn{3}{c}{LexTALE} & \multicolumn{3}{c}{Michigan} \\ \cmidrule{2-7}
 & \multicolumn{1}{c}{EyeScore} & \multicolumn{2}{c|}{$\text{EyeScore}^{\mathrm{db}}$} & \multicolumn{1}{c}{EyeScore} & \multicolumn{2}{c}{$\text{EyeScore}^{\mathrm{db}}$} \\ \cmidrule{2-7}
\textbf{Features} &  & \textit{Seen L1} & \textit{New L1} &  & \textit{Seen L1} & \textit{New L1} \\
\midrule \cmidrule{1-7}
WPM & $\phantom{-}0.01$ & $-0.01$ & $\phantom{-}0.31$ & $-1.51\phantom{^{*}}$ & $-1.46$ & $-1.22$ \\
\midrule
Avg. Fix. & $-0.30$ & $-0.08$ & $\phantom{-}0.26$ & $-1.09\phantom{^{*}}$ & $-0.83$ & $-0.51$ \\
S-Clusters & $-0.20$ & $-0.06$ & $\phantom{-}0.20$ & $-0.91\phantom{^{*}}$ & $-0.73$ & $-0.50$ \\
WP-Coefs & $-0.39$ & $-0.02$ & $\phantom{-}0.08$ & $-0.85^{*}$ & $-0.48$ & $-0.37$ \\ \bottomrule
\end{tabular}%
}
\captionof{table}{\textbf{L1-English Proximity Bias in EyeScore and EyeScore\textsuperscript{db}} in the \textbf{Any Text} regime, for the \textbf{OneStopL2 information seeking} reading regime.}
\label{tab:lang-bias-triplet-coef-two_step-hunting}
\endgroup

\vspace{0.4em}
}]

\twocolumn[{%
\subsection{EyeScore$^\mathrm{\textbf{db}}$
Correlations with Standard Proficiency Tests}
\label{sec:bias-hurt-corr}
\vspace{0.4em}

\centering
\setlength{\aboverulesep}{0pt}
\setlength{\belowrulesep}{0pt}
\renewcommand{\arraystretch}{1.25}
\resizebox{\textwidth}{!}{
\begin{tabular}{lcccc|cccc|cccc|cccc}
\toprule
\multicolumn{1}{c}{} & \multicolumn{8}{c}{\textbf{MECO L2}} & \multicolumn{8}{c}{\textbf{OneStopL2}} \\
\multicolumn{1}{c}{} & \multicolumn{4}{c}{LexTALE} & \multicolumn{4}{c|}{Composite} & \multicolumn{4}{c}{LexTALE} & \multicolumn{4}{c}{Michigan} \\ \cmidrule{2-17}
\multicolumn{1}{c}{} & \multicolumn{2}{c}{Fixed} & \multicolumn{2}{c!{\vrule height 0.7\ht\strutbox depth \dp\strutbox}}{Any} & \multicolumn{2}{c}{Fixed} & \multicolumn{2}{c|}{Any} & \multicolumn{2}{c}{Fixed} & \multicolumn{2}{c!{\vrule height 0.7\ht\strutbox depth \dp\strutbox}}{Any} & \multicolumn{2}{c}{Fixed} & \multicolumn{2}{c}{Any} \\ \cmidrule{2-17}
\textbf{Features} & Seen L1 & New L1 & Seen L1 & New L1 & Seen L1 & New L1 & Seen L1 & New L1 & Seen L1 & New L1 & Seen L1 & New L1 & Seen L1 & New L1 & Seen L1 & New L1 \\
\midrule
WPM & -0.015$^{***}$ & -0.002 & -0.015$^{***}$ & -0.002 & -0.002 & +0.015$^{**}$ & -0.002 & +0.015$^{***}$ & -0.030$^{**}$ & -0.030$^{*}$ & -0.017 & -0.011 & -0.021 & -0.024 & -0.005 & -0.005 \\
\midrule
Avg. Fix. & -0.013$^{***}$ & -0.002 & -0.013$^{***}$ & -0.003 & -0.002 & +0.011$^{**}$ & -0.001 & +0.011$^{**}$ & -0.011 & -0.013 & -0.014 & -0.016 & -0.007 & -0.007 & -0.002 & -0.011 \\
S-Clusters & -0.012$^{***}$ & -0.002 & -0.011$^{**}$ & -0.001 & -0.002 & +0.011$^{**}$ & -0.001 & +0.011$^{**}$ & -0.012 & -0.016 & -0.014 & -0.015 & -0.008 & -0.008 & -0.002 & -0.009 \\
WP-Coefs & -0.015$^{***}$ & -0.006 & -0.016$^{***}$ & -0.007 & +0.001 & +0.015$^{***}$ & +0.002 & +0.015$^{**}$ & -0.010 & -0.020 & -0.020 & -0.019 & +0.008 & -0.010 & -0.002 & -0.017 \\
\midrule
Transitions & +0.014$^{***}$ & +0.017$^{**}$ & - & - & +0.016$^{***}$ & +0.016$^{**}$ & - & - & +0.016 & +0.014 & - & - & +0.005 & -0.003 & - & - \\
Word Fix. & -0.016$^{***}$ & -0.006 & - & - & +0.000 & +0.015$^{**}$ & - & - & -0.017 & -0.018 & - & - & -0.012 & -0.013 & - & - \\
\bottomrule
\end{tabular}
}
\captionof{table}{\textbf{The effect of EyeScore L1 debiasing on correlations with standard proficiency tests}. Presented is $\Delta r_{\mathrm{db}}$, the Pearson $r$ of EyeScore\textsuperscript{db} with each standard English proficiency test minus the Pearson $r$ of EyeScore with the same test. Seen L1 / New L1: did the data for computing the debiased score include participants from the same L1 as the test participant. Statistically significant differences  using a bootstrap test are marked with `$^{*}$' $p<0.05$, `$^{**}$' $p<0.01$, `$^{***}$' $p<0.001$.
}
\label{tab:eyescore_seen_unseen-debiased-lextale-diff}

\vspace{0.4em}
}]

\clearpage

\section{Test Reliability}
\label{sec:app-test-reliability}

\subsection{Comparison of EyeScore$^{\mathrm{\textbf{items}}}$ and EyeScore}
\label{sec:eyescoreitems-eyescore-comparison}

\begin{table}[ht!]
\centering
\small
\begin{tabular}{lcc|cc}
\toprule
\multirow{2}{*}{\textbf{Features}} & \multicolumn{2}{c|}{\textbf{MECO L2}} & \multicolumn{2}{c}{\textbf{OneStopL2}} \\
 & Fixed & Any & Fixed & Any \\
\midrule
WPM & 1.00 & 1.00 & 0.99 & 1.00 \\
\midrule
Avg. Fix. & 0.99 & 0.99 & 0.98 & 0.98 \\
S-Clusters & 0.99 & 0.99 & 0.98 & 0.98 \\
WP-Coefs & 0.95 & 0.93 & 0.96 & 0.97 \\
\midrule
Transitions & 0.99 & - & 0.98 & - \\
Word Fix. & 1.00 & - & 1.00 & - \\
\bottomrule
\end{tabular}
\caption{Pearson $r$ correlation between \textbf{EyeScore$^{\mathrm{\textbf{items}}}$} and \textbf{EyeScore}.}
\label{tab:original_vs_per_item_eyescore}
\end{table}

\begin{table}[ht!]
\centering
\resizebox{\columnwidth}{!}{
\begin{tabular}{lcccc|cccc}
\toprule
 \multicolumn{1}{c}{} & \multicolumn{4}{c|}{\textbf{MECO L2}} & \multicolumn{4}{c}{\textbf{OneStopL2}} \\
 \multicolumn{1}{c}{} & \multicolumn{2}{c}{LexTALE} & \multicolumn{2}{c|}{Composite} & \multicolumn{2}{c}{LexTALE} & \multicolumn{2}{c}{Michigan} \\
\cmidrule{2-9}
\textbf{Features} & Fixed & Any & Fixed & Any & Fixed & Any & Fixed & Any \\
\midrule
WPM & +0.01 & +0.01 & +0.01 & +0.01 & 0.00 & 0.00 & -0.01 & 0.00 \\
\midrule
Avg. Fix. & +0.01 & +0.01 & +0.01 & +0.01 & +0.02 & +0.03 & +0.02 & +0.03 \\
S-Clusters & +0.02 & +0.02 & +0.02 & +0.02 & +0.03 & +0.03 & +0.03 & +0.03 \\
WP-Coefs & -0.01 & -0.01 & -0.02 & -0.02 & +0.04 & +0.04 & 0.00 & +0.01 \\
\midrule
Transitions & 0.00 & - & +0.02 & - & +0.02 & - & +0.01 & - \\
Word Fix. & 0.00 & - & 0.00 & - & 0.00 & - & 0.00 & - \\
\bottomrule
\end{tabular}
}
\caption{\textbf{EyeScore$^{\mathrm{\textbf{items}}}$} Pearson $r$ correlation with standard proficiency tests.  Presented is $\Delta r_{\mathrm{items}}$, the Pearson $r$ of EyeScore$^{\mathrm{items}}$ with each standard English proficiency test minus the Pearson $r$ of EyeScore with the same test. }
\label{tab:evaluation-eyescore-delta}
\end{table}

\subsection{Internal Consistency Evaluation Procedure}
\label{sec:app-eval-setting-per-item}

As described in \Cref{sec:internal-consistency}, the item-level analysis requires a separate EyeScore for each passage read by each participant. We therefore apply the EyeScore procedure to passage-level feature vectors, such that each participant is represented by one vector for each recorded passage. The way these vectors are used to compute EyeScore differs between the Fixed Text and Any Text regimes.

In the \textbf{Fixed Text regime}, when calculating each item's EyeScore, we are only interested in feature vectors extracted for the same item across participants. This resembles the original setup, where each participant has only one vector representing them, and we repeat this process for each item. 

In the \textbf{Any Text regime}, when calculating each item's EyeScore, each training participant contributes multiple passage-level feature vectors. We perform the normalization over all such vectors, effectively normalizing them relative to all passage-level vectors of L2 participants in train, and construct the L1 prototype analogously as the average L1 participant over passages.

\subsection{Split-Half Reliability Evaluation Procedure}
\label{sec:app-eval-setting-spltihalf}

In MECO L2, as mentioned in \Cref{sec:eval-setting}, we divide the 12 passages into two parts with 6 passages each. To preserve this balance in the split-half reliability analysis, we randomly split each of the two parts separately. Since not all participants have eye movement data for all passages, a split may result in a participant appearing in only one of the halves. Thus, for each split, we include only participants with eye movement data for at least one passage in each half. On average, this filtering removes 5 of the 1,106 participants per split.

As mentioned in \Cref{sec:eval-setting}, in OneStopL2 and OneStop, there are 6 participant subgroups, each group reading its own set of passages. Thus, for each split of the split-half reliability analysis, we randomly split each subgroup's set of passages separately and combine the corresponding halves across subgroups.

\subsection{Reliability in the Fixed Text Regime and in Information Seeking}
\label{sec:app-reliability-eyescore-extended}

\begin{table}[ht!]
\centering
\resizebox{\columnwidth}{!}{
\begin{tabular}{lccc|ccc}
\toprule
 \multicolumn{1}{c}{} & \multicolumn{3}{c|}{\textbf{MECO L2}} & \multicolumn{3}{c}{\textbf{OneStopL2}} \\
\textbf{Features} & Cronbach's $\alpha$ & SEM & Split-Half $r$ & Cronbach's $\alpha$ & SEM & Split-Half $r$ \\
\midrule
WPM & 0.98 & 0.13 & 0.93 & 0.99 & 0.09 & 0.98 \\
\midrule
Avg. Fix. & 0.98 & 0.08 & 0.92 & 0.99 & 0.04 & 0.99 \\
S-Clusters & 0.98 & 0.04 & 0.94 & 1.00 & 0.02 & 0.99 \\
WP-Coefs & 0.92 & 0.06 & 0.79 & 0.98 & 0.03 & 0.94 \\
\midrule
Transitions & 0.97 & 0.01 & 0.91 & 0.99 & 0.01 & 0.99 \\
Word Fix. & 0.98 & 0.04 & 0.95 & 1.00 & 0.02 & 0.99 \\
\bottomrule
\end{tabular}
}
\caption{\textbf{Reliability of EyeScore} in the \textbf{Fixed Text} regime.}
\label{tab:eyescore-reliability-fixed-paragraph}
\end{table}

\begin{table}[ht!]
\centering
\resizebox{\columnwidth}{!}{
\begin{tabular}{lccc|ccc}
\toprule
 \multicolumn{1}{c}{} & \multicolumn{3}{c|}{\textbf{Fixed Text}} & \multicolumn{3}{c}{\textbf{Any Text}} \\
\textbf{Features} & Cronbach's $\alpha$ & SEM & Split-Half $r$ & Cronbach's $\alpha$ & SEM & Split-Half $r$ \\
\midrule
WPM & 0.98 & 0.13 & 0.97 & 0.98 & 0.09 & 0.97 \\
\midrule
Avg. Fix. & 0.99 & 0.05 & 0.98 & 0.99 & 0.05 & 0.98 \\
S-Clusters & 0.99 & 0.03 & 0.98 & 0.99 & 0.03 & 0.98 \\
WP-Coefs & 0.98 & 0.04 & 0.91 & 0.99 & 0.03 & 0.94 \\
\midrule
Transitions & 0.99 & 0.01 & 0.98 & - & - & - \\
Word Fix. & 0.99 & 0.03 & 0.99 & - & - & - \\
\bottomrule
\end{tabular}
}
\caption{\textbf{Reliability of EyeScore} in the \textbf{OneStopL2 information seeking} reading regime.}
\label{tab:eyescore-reliability-regimes-paragraph-information-seeking}
\end{table}

\subsection{Reliability of Standard English Proficiency Tests}
\label{sec:app-reliability-comparison-other}

\textbf{Michigan Placement Test (OneStopL2)} Split-half Pearson $r$ was calculated over 20 splits, with a balanced number of items from each test section across the two halves.
The resulting Cronbach's $\alpha$ of the listening part is $0.74$, for the grammar section it is $0.84$, for the vocabulary section it is $0.82$, and for the reading section it is $0.78$. The stratified Cronbach's $\alpha$ \citep{cronbach1965alpha} overall is $0.91$, and the average Pearson $r$ for split-half reliability is $0.92$. In the OneStopL2 information seeking regime, the reliability scores for the section are $0.75$ for listening, $0.85$ for grammar, $0.83$ for vocabulary, and $0.78$ for reading. The overall stratified Cronbach's $\alpha$ of $0.95$ and the split-half Pearson $r$ is $0.96$. 

\textbf{IELTS} 
The IELTS test comprises four sections: listening, reading (in one of two variants, `academic' or `general training'), speaking, and writing. 
\citet{IELTS2026} report average Cronbach's $\alpha$ values of $0.90$ for the listening section, $0.92$ for academic reading, $0.91$ for general training reading, as well as inter-rater reliability estimates of $0.90$ for speaking and $0.92$ for writing. 

\textbf{TOEFL-iBT} \citet{manna2025toefl} report Cronbach's $\alpha$ of $0.94$ for the speaking section and $0.87$ for the writing section (stratified). The reading and listening parts of the test are evaluated using a reliability estimate based on item response theory (IRT)  \citep{kolen1996conditional}, with a reliability score of $0.86$ for the reading section and $0.88$ for the listening section. The reported overall reliability score is $0.90$.

\end{document}